\documentclass[conference]{IEEEtran}

\usepackage[utf8]{inputenc}
\usepackage[T1]{fontenc}

\IEEEoverridecommandlockouts

\usepackage{amsmath,amsfonts,bm}

\def\eqref#1{equation~\ref{#1}}

\def\1{\bm{1}}

\DeclareMathAlphabet{\mathsfit}{\encodingdefault}{\sfdefault}{m}{sl}
\SetMathAlphabet{\mathsfit}{bold}{\encodingdefault}{\sfdefault}{bx}{n}

\usepackage{hyperref}
\usepackage{natbib}
\usepackage{url}

\usepackage[utf8]{inputenc} 
\usepackage[T1]{fontenc}    
\usepackage{booktabs}       
\usepackage{amsfonts}       
\usepackage{nicefrac}       
\usepackage{microtype}      
\usepackage{xcolor}         
\usepackage{amsmath}
\usepackage{graphicx}
\usepackage{xspace}
\newcommand{\bench}{{MM-ReportBench}\xspace}
\newcommand{\TJudge}{{T-Judge}\xspace}
\newcommand{\VJudge}{{V-Judge}\xspace}

\def\BibTeX{{\rm B\kern-.05em{\sc i\kern-.025em b}\kern-.08em
    T\kern-.1667em\lower.7ex\hbox{E}\kern-.125emX}}
\begin{document}

\title{MM-ReportBench: A Comprehensive Benchmark for Multimodal Report Generation from Tables\\
}

\author{\IEEEauthorblockN{Teng Lin}
\IEEEauthorblockA{\textit{DSA Thrust} \\
\textit{HKUST(GZ)}\\
Guangzhou, China \\
tlin280@connect.hkust-gz.edu.cn}
\and
\IEEEauthorblockN{Nan Tang}
\IEEEauthorblockA{\textit{DSA Thrust} \\
\textit{HKUST(GZ)}\\
Guangzhou, China \\
nantang@hkust-gz.edu.cn}
}

\maketitle

\begin{abstract}
The ability to generate multimodal reports integrating fluent narratives with faithful visualizations from tabular data, has become a cornerstone capability for modern vision‑language models in practical domains such as finance, healthcare, and policy making. However, existing benchmarks are limited by single-modality focus, simplistic metrics, novelty evaluation, and visualization quality. We address these limitations with \bench, a multimodal report evaluation benchmark covering 386 tasks across 185 real-world tables. Each task is paired with an expert-curated reference report featuring standardized chapter structures and verifiable key insight points, constructed through a hybrid pipeline of multimodal reports generation based on Monte Carlo Tree Search and rigorous human quality control. For robust evaluation, we design a dual‑judge evaluation framework: Text Judge Model assesses textual quality along structural completeness, grounded insight verification with numerical fact-checking against source tables, and novelty analysis that distinguishes genuine analytical discovery from trivial paraphrasing; Visualization Judge Model evaluates generated visualizations across data fidelity, expressiveness, aesthetics, and technical correctness. They also evaluate cross-modal consistency. Extensive experiments on 20 state-of-the-art models reveal critical weaknesses: even the strongest model achieves only 68.9 overall score, with numerical hallucination affecting 38.5\% of reports and genuinely novel insights remaining rare. Our ablation studies further demonstrate that single-judge evaluation consistently inflates scores by 8–12 points compared to dual-judge assessment, confirming that rigorous and reliable benchmarking demands multi‑faceted, complementary evaluation probes. 
\end{abstract}

\section{Introduction}
\label{sec:introduction}

The rapid evolution of large language models (LLMs) and large vision-language models (LVLMs) has fundamentally reshaped how organizations interact with data. In real-world business analytics, scientific discovery, and decision-making, a critical emerging capability is \textit{multimodal report generation}: the automated production of professional reports that synthesize textual analysis, quantitative insights, and visual charts directly from raw structured tables. Unlike conventional table question answering or simple summarization, this task demands that models not only extract and aggregate numerical facts but also structure coherent narratives, ground claims in source evidence, generate informative visualizations, and deliver actionable conclusions while maintaining factual fidelity and logical coherence~\cite{huang2026mmdeepresearch, liu2026huntinstead, lin2026montecarlotreesearch,lin2026cominsight,lin2025lightkggsimpleefficientknowledge, lin2025Simplifying}. As enterprises increasingly deploy AI agents for end-to-end data analysis workflows, the ability to reliably evaluate such multimodal report generation systems has become both an urgent practical need and a significant research challenge.
 
Despite substantial progress in table reasoning benchmarks, existing works exhibit inadequate modes. First, current benchmarks predominantly focus on single modalities in isolation, like table question answering~\cite{zhong2017seq2sql,nan2022fetaqa, Chen2020TabFact, parikh2020totto,lin2025structured,lin-etal-2025-mebench}, while recent efforts toward open-ended report writing such as T2R-Bench~\cite{zhang2025t2rbench} and DDR-Bench~\cite{liu2026huntinstead} remain confined to textual outputs and do not assess the generation of visualizations essential to professional business reports. Second, the LLM-as-a-judge paradigm, though widely adopted for scalability, suffers from fundamental vulnerabilities: It is demonstrated that reasoning judges can drive policy models toward adversarial output strategies that achieve high judge rewards while violating task intent, and that such adversarial outputs generalize to deceive strong evaluators, underscoring the need for evaluation frameworks resistant to reward hacking~\cite{liu2026examining}. Third, existing benchmarks lack standardized reference standards with verifiable ground truth, MMDR-Bench~\cite{huang2026mmdeepresearch} proposes multi-stage evaluation with visual evidence fidelity checks but depends on retrieval-based citation verification unsuitable for table-grounded settings, while some works establishes comprehensive visualization assessment yet evaluates charts in isolation without connection to textual reports and source tables. Across these efforts, a unified benchmark that integrates structured tabular data, multimodal report generation, and rigorous multi-aspect evaluation with verifiable reference standards remains absent.

To bridge this gap, we introduce \textbf{\bench}, a comprehensive benchmark for table-to-multimodal report generation with a comprehensive dual-judge evaluation suite. Our design is directly informed by the lessons of prior work: we ground all factuality checks in source tables rather than external retrieval to mitigate evaluation manipulation; we split evaluation into specialized judges to prevent the halo effects where strong text masks poor visuals or vice versa; and we incorporate explicit novelty assessment to reward genuine analytical discovery over paraphrasing.

Central to our benchmark is a \textbf{dual-judge evaluation framework}, which we adopt to overcome a critical pitfall of prior evaluation protocols: monolithic scoring systems that aggregate textual and visual quality into a single metric often conflate heterogeneous failure modes, mask modality-specific errors, and produce inflated overall scores, for example, a poorly grounded narrative might be camouflaged by aesthetically pleasing charts, or vice versa. By decoupling the assessment of generated text and visualizations, our framework offers two distinct advantages: (a) it enables fine-grained diagnostic analysis, pinpointing exactly whether a model struggles with factual grounding, chart fidelity, or creative insight; and (b) it prevents cross-modality compensation, ensuring that excellence in one modality cannot artificially boost the perceived performance of the other. Concretely, the framework comprises two independent judges: \textbf{\TJudge} evaluates the textual report along three axes, including Structural Completeness, Grounded Insight Verification and Novelty Analysis;\textbf{\VJudge} assesses generated visualizations across four dimensions including Data Fidelity, Expressiveness, Aesthetics and Technical Correctness. And they assess the \textbf{cross-modal consistency} together. This modular design not only mirrors the real-world expectation that human analysts critique reports and charts separately, but also provides a principled foundation for our subsequent ablation studies, which confirm that a single judge inflates scores by 8-12 points relative to the dual-judge setup, vindicating the necessity of this separation.

We conduct extensive experiments on over 20 state-of-the-art models, including text-only LLMs, vision-language models, code-augmented multimodal systems, and specialized deep research agents reveal critical weaknesses across the board. The contributions of this work are:

\begin{itemize}
    \item \textbf{We propose the first comprehensive benchmark for table-to-multimodal report generation}, with 386 tasks across six business domains and expert-curated reference reports with verifiable key insight points.
    \item We design a \textbf{dual-judge evaluation} framework combining specialized \TJudge and \VJudge modules that jointly measure textual and visual quality through interpretable, dimension-level scoring.
    \item We propose a rigorous reference construction pipeline using \textbf{MCTS with multi-dimensional rewards} and multi-stage human quality control.
    \item \textbf{Comprehensive empirical findings} exposing critical failure modes in current models and actionable insights for advancing reliable multimodal report generation.
\end{itemize}

\section{Related Work}
\label{sec:related}

\textbf{Table Reasoning and Report Generation Benchmarks.}
Early benchmarks for table understanding focused on narrow, well-defined tasks such as table-based question answering~\cite{zhong2017seq2sql, nan2022fetaqa, Chen2020TabFact, pasupat2015compositional, wu2025tablebench, lin2025srag,lin-etal-2025-mebench, lin2026docsageinformationstructuringagent,lin2025structured}, table-to-text generation~\cite{parikh2020totto, cheng2022hitab,lin2026structurethenquery,lin2026annoretrieveefficientstructuredretrieval}, and text-to-SQL parsing~\cite{yu2018spider}. While these datasets enabled significant progress in basic table comprehension, they largely ignore the complex, open-ended requirements of real-world business analytics. More recent work has started to address this gap. T2R‑Bench~\cite{zhang2025t2rbench} formalizes the table‑to‑report task with industrial tables and evaluates numerical accuracy, coverage, and writing quality, while DDR‑Bench~\cite{liu2026huntinstead} advances autonomous data exploration.
However, both frameworks remain confined to textual outputs and do not assess the generation of visualizations, which are essential components of professional reports. 

\textbf{LLM-as-a-Judge and Multimodal Evaluation.}
The LLM‑as‑a‑judge paradigm is vulnerable to reward hacking and adversarial outputs: It is shown that reasoning‑judge‑trained policies learn systematic strategies that deceive both training and downstream judges, including GPT‑4.1 on Arena‑Hard ~\cite{liu2026examining}. Recent reasoning judges such as JudgeLRM~\cite{chen2025judgelrm} and RM‑R1~\cite{chen2025rm} improve static benchmark scores but have not been rigorously tested in policy training. For visualization assessment, VisEval~\cite{chen2024viseval} and VIS‑Shepherd~\cite{pan2025vis} employ LLMs to judge charts within NL2VIS pipelines, while MMDR‑Bench~\cite{huang2026mmdeepresearch} proposes multi‑stage evaluation with visual evidence fidelity checks for deep research reports, and VisJudge‑Bench~\cite{xie2025visjudge} builds a specialized model for chart aesthetics and quality. Despite its comprehensiveness, MMDR-Bench targets research reports over image-text bundles rather than structured tables, and its evaluation heavily depends on retrieval-based citation verification, which is not directly applicable to table-grounded reports. Our dual-judge framework verifies all claims directly against source tables to reduce evaluation fragility and judge-manipulability.

\section{\bench}
\label{sec:system}


\subsection{Task Formulation}
\label{sec:task}

The multimodal report generation task is defined as follows: Given a set of one or more structured tables $\mathcal{T} = \{T_1, T_2, \dots, T_n\}$ and a natural language instruction $q$ that specifies the analytical focus (e.g., ``Employment trends across U.S. sectors since 2006''), a model must produce a multimodal report $\mathcal{R} = (R_{\text{text}}, R_{\text{vis}})$. The textual component $R_{\text{text}}$ is an article-length analytical report containing structured chapters (e.g., Overview, Data Overview, Core Analysis, Conclusions \& Recommendations), while the visual component $R_{\text{vis}}$ consists of one or more generated charts (bar charts, line plots, pie charts, scatter plots, etc.) that are embedded within the text and directly support the narrative with underlying table evidence. The report is evaluated against a human-authored reference, which specifies an expected chapter structure and a set of key insight points $K = \{k_1,\dots,k_M\}$ that capture the essential analytical findings derivable from the tables. Unlike prior benchmarks that evaluate text and visuals independently or ignore visual output entirely, \bench treats the report as an integrated multimodal artifact, requiring coherence across modalities.

\subsection{Dataset Construction}
\label{sec:dataset}

\textbf{(1) Table Collection and Domain Coverage.}
We collect 185 real-world business tables from publicly available sources including municipal open data platforms, industry association reports, corporate annual disclosures, and open-source tabular datasets, following the protocol of T2R-Bench. Tables span six high-impact industrial domains: finance, manufacturing, healthcare, education, retail, and IT operations. We specifically include 38 extremely large tables (over 50,000 cells, 20.5\% of the dataset) and tables with complex header structures (32.4\%) to test models' capabilities in processing massive-scale and hierarchical structured data. All tables undergo manual review to remove personally identifiable information and verify sufficient quantitative depth for meaningful analysis and visualization. The final collection contains bilingual tables (Chinese and English), reflecting the demands of global business analytics. Table~\ref{tab:dataset_stats} summarizes the key statistics of the \bench dataset.

\begin{figure*}[t]
\centering
\includegraphics[width=0.9\linewidth]{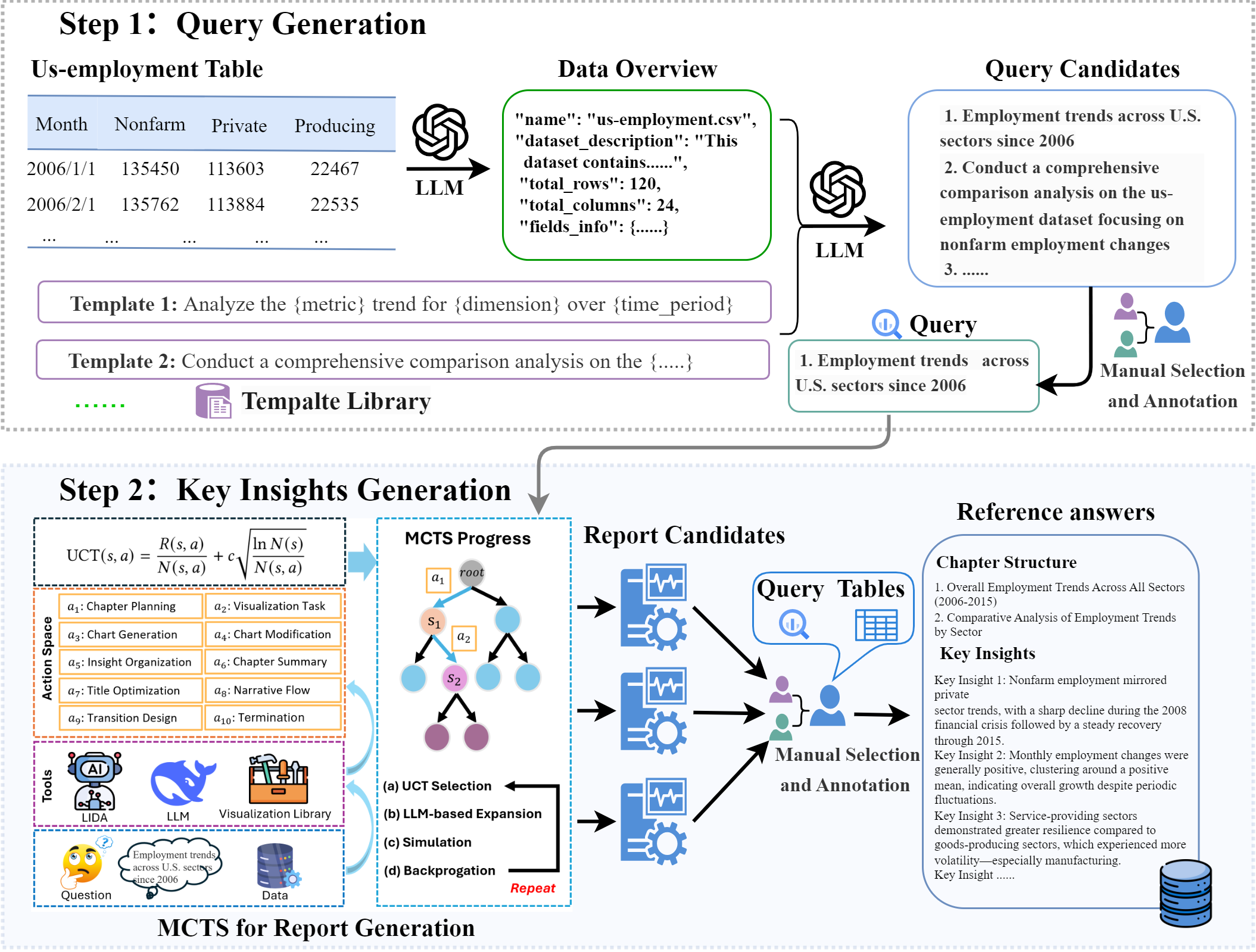}
\caption{An overview of the construction pipeline for \bench.}
\label{fig:dataconstruction}
\end{figure*}

\begin{table}[t]
\centering
\caption{Key statistics of MM-ReportBench.}
\label{tab:dataset_stats}
\footnotesize
\begin{minipage}[t]{0.48\textwidth}
\centering
\begin{tabular}{p{4.5cm} r}
\toprule
\textbf{Property} & \textbf{Value} \\
\midrule
Number of tables & 185 \\
Number of domains & 6 \\
Number of tasks & 386 \\
Average tasks per table & 2.09 \\
Average tables per multi-table task & 4.2 \\
Average rows per table & 285 \\
\bottomrule
\end{tabular}
\end{minipage}
\hfill
\begin{minipage}[t]{0.48\textwidth}
\centering
\begin{tabular}{p{4.5cm} r}
\toprule
\textbf{Property} & \textbf{Value} \\
\midrule
Average cells per table & 21200 \\
Extremely large tables ($> 50$K cells) & 38 \\
Complex header tables & 60 \\
Average reference keypoints per task & 7.75 \\
Total reference keypoints & 2992\\
\bottomrule
\end{tabular}
\end{minipage}
\end{table}

\textbf{(2) Query Generation and Selection.}
We design a template-based self-instruct approach for analytical query generation. Ten templates covering common business reporting scenarios (trend analysis, comparative evaluation, ranking, anomaly detection, composition analysis, correlation exploration, forecasting, segment analysis, year-over-year change, cross-dimensional interaction) are populated with table-specific metadata and expanded into 3–5 concrete query candidates by GPT-4o. Two independent domain-expert and one senior expert annotators evaluate candidates based on answerability (fully solvable from provided tables without external knowledge), analytical focus (targeting a single coherent narrative thread), and complementarity (distinct, non-overlapping analytical dimensions for multiple queries per table). This yields 386 high-quality queries. The pipeline is shown in Figure~\ref{fig:dataconstruction} step 1 Query Generation.

\textbf{(3) Reference Report Generation via Monte Carlo Tree Search.}
A central challenge in constructing a report-generation benchmark is obtaining high-quality reference reports that serve as stable evaluation targets. Human-authored reports are expensive to scale, while LLM-generated reports often suffer from hallucinations and verifiability issues. To balance quality, scalability, and diversity, we design a pipeline which reference report candidates are first generated through a Monte Carlo Tree Search (MCTS)~\cite{mcts} procedure guided by a multi-dimensional reward function, and then curated and refined by human experts, which is demonstrated in Figure~\ref{fig:dataconstruction} MCTS for Report Generation step.

\textit{MCTS for Report Generation.} We formulate report generation as sequential decision-making over an action space with two stages: Analysis (chapter planning, visualization task identification, chart generation, chart modification, insight organization, chapter summarization) and Refinement (title optimization, narrative flow improvement, and transition design). From an empty root node, the tree is expanded through iterative rollouts with action selection using the UCT~\cite{uct} criterion:

\begin{equation}
\text{UCT}(s, a) = \frac{R(s, a)}{N(s, a)} + c \sqrt{\frac{\ln N(s)}{N(s, a)}}
\end{equation}

where $R(s, a)$ is the cumulative reward obtained after taking action $a$ in state $s$, $N(s, a)$ is the visit count for $(s, a)$, $N(s)$ is the total visit count for state $s$, and $c$ is the exploration constant. We perform $K=8$ rollouts per query, producing $K$ complete report candidates.

\textit{Multi-Dimensional Report Reward Function.}
To guide the MCTS search toward high-quality, reference-worthy reports, we design a reward function $R(\mathcal{R})$ that evaluates a completed report along four dimensions, inspired by the evaluation criteria proposed in this work:

\begin{equation}
R(\mathcal{R}) =  R_{\text{rep}}(\mathcal{R}) + R_{\text{pres}}(\mathcal{R}) +  R_{\text{aes}}(\mathcal{R}) + R_{\text{nar}}(\mathcal{R})
\end{equation}

where \(R_{\text{rep}}\) captures data representation accuracy (combining numerical accuracy and information coverage metrics), \(R_{\text{pres}}\) assesses structural clarity and logical organization via LLM judge, \(R_{\text{aes}}\) evaluates visual chart quality, and \(R_{\text{nar}}\) assesses storytelling coherence and analytical depth (rewarding causal reasoning and actionable recommendations). Then we select the top $K=3$ candidates with the highest terminal rewards to pass to the human curation stage. For more detailed information of MCTS for report generation, please refer to the appendix~\ref{subsec:MCTS}.

\textit{Human Curation and Keypoint Annotation.}
The MCTS-generated report candidates per query undergo two-stage human curation. Two Annotators select the best candidate and collaboratively refine it through structural standardization, factual correction, and keypoint extraction (5–10 concise factual statements per report). Disagreements are resolved by a senior expert. The final output per task is a human-curated reference report with a defined chapter structure and keypoints. To further ensure the quality, we implement a three-stage verification pipeline: (1) factual verification where every numerical claim in each keypoint is cross-checked against raw table data by executing corresponding SQL queries; (2) necessity and exclusivity review where a dedicated quality-control team validates that each keypoint is necessary for the report's analytical goals and exclusively derivable from provided tables, removing trivial keypoints; and (3) we implement a reproducibility audit, where 50 randomly selected tasks are independently re-derived by three separate annotators using the same MCTS-assisted pipeline, achieving a Fleiss' \(\kappa\) of 0.85 with the original references and confirming the reliability of our reference construction process.





\subsection{Dual-Judge Evaluation Framework}
\label{sec:evaluation}

\begin{figure*}[t]
\centering
\includegraphics[width=0.9\linewidth]{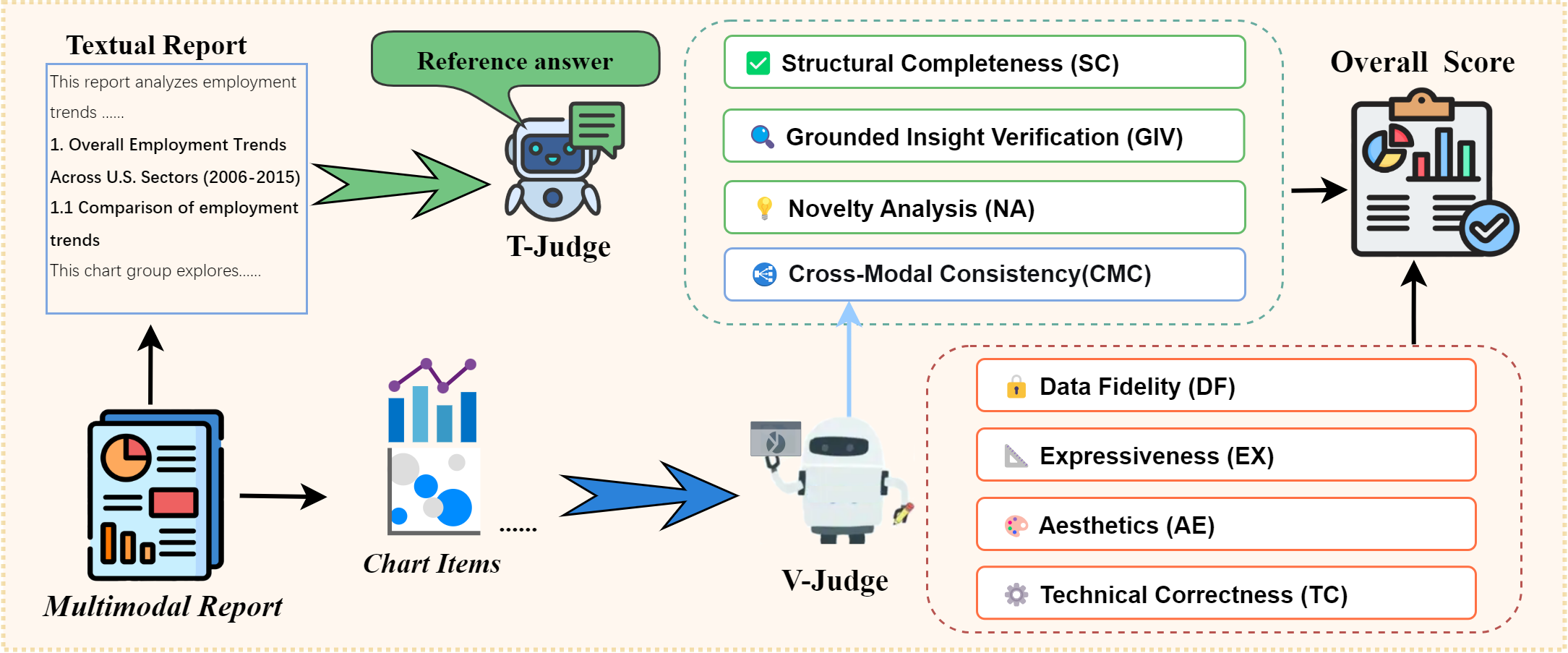}
\caption{The Dual-Judge Evaluation Framework of \bench. CMC verifies whether textual references to charts and chart-supported narratives are mutually consistent.}
\label{fig:evaluate}
\end{figure*}

As shown in Figure~\ref{fig:evaluate}, we design a dual-judge evaluation suite. Both modules operate with interpretable, dimension-level scoring.

\textbf{(1) \TJudge.}
\label{sec:report_judge}
\TJudge evaluates \(R_{\text{text}}\) along three orthogonal dimensions:

\textit{(a) Structural Completeness (SC).}
This dimension measures how well the generated report's chapter organization, section flow, and topical coverage align with the reference chapter structure. We compute heading overlap score (F1 against reference headings with semantic equivalence via embedding-based matching, threshold 0.85) and section content coverage (\TJudge assesses whether matched sections cover the same analytical sub-dimensions). The final SC score is \(0.3\cdot \text{SC}_{\text{head}} + 0.7\cdot \text{SC}_{\text{content}}\), emphasizing content-level alignment over surface formatting.

\textit{(b) Grounded Insight Verification (GIV).}
This dimension assesses factual faithfulness through two sub-metrics: (b1) \textit{Numerical Accuracy} (\(\text{GIV}_{\text{num}}\)): We adapt the NAC protocol from T2R-Bench. First, we extract numerical sentence clusters, generate verification questions with the claimed values as target answers, and have three specialized code-generation LLMs independently answer by executing SQL queries on source tables. A claim is verified if at least two of three models agree within a \(1\%\) tolerance margin. (b2) \textit{Keypoint Coverage} (\(\text{GIV}_{\text{cov}}\)): We measure semantic alignment between generated report sentences and reference keypoints via normalized mutual information:
    \begin{equation}
        \text{GIV}_{\text{cov}} = \frac{\sum_{i=1}^{M}\sum_{j=1}^{N}P(k_i,s_j)\log\frac{P(k_i,s_j)}{P(k_i)P(s_j)}}{-\sum_{i=1}^{M}P(k_i)\log P(k_i)}
    \end{equation}
where probabilities are derived from a BERTScore-based similarity matrix, which provides a computable, reproducible, and paraphrase‑robust metric. More detailed information can be found in Appendix~\ref{subsec:BERTScore}.     
The composite GIV score is the average of \(\text{GIV}_{\text{num}} \) and \(\text{GIV}_{\text{cov}}\).

\textit{(c) Novelty Analysis (NA).}
To address the challenge that prior benchmarks cannot distinguish between paraphrasing and genuine insight discovery, we identify novel claims: sentence clusters with maximum cosine similarity below \(\tau_{\text{novel}} = 0.6\) to any reference keypoint. These claims are evaluated by \TJudge along three sub-dimensions: non-triviality, actionability, and correctness (factual consistency with source tables). The overall NA score incorporates a penalty factor \(\min(1, \alpha / |\mathcal{N}| )\) with \(\alpha = 5\) to prevent models from receiving high scores through large quantities of low-quality novel claims, rewarding focused, high-quality insights over volume.

\textit{(d) Cross-Modal Consistency (CMC).} It includes two sub-metrics:
(d1) Cross-Modal Reference Consistency (CMRC). For each sentence that explicitly references a chart, T-Judge checks whether the referenced chart exists. CMRC = correctly grounded chart-referencing sentences / all chart-referencing sentences. (d2) Narrative Alignment (NRA). For each generated chart, \textbf{V-Judge} checks whether the chart is explained by the text, whether the described trend/values/labels match the chart, and whether the chart supports the section’s analytical claim. NRA = correctly explained charts / all charts.
The CMC value is the average of CMRC and NRA.
The final \TJudge score is the average of \text{SC}, \text{GIV}, \text{NA} and \text{CMC}.

\textbf{(2) \VJudge.}
\label{sec:vis_judge}
\VJudge evaluates \(R_{\text{vis}}\) on four dimensions for chart-specific verification:

\textit{(a) Data Fidelity (DF).}
This dimension verifies plotted data against underlying tables through a two-step process: chart data extraction using a chart-to-table model and ground-truth comparison via SQL-generated expected data. Charts with more than \(5\%\) data points deviating beyond \(1\%\) tolerance (absolute) or 2 percentage points (proportions) are flagged as fidelity failures. 

\textit{(b) Expressiveness (EX).}
This dimension evaluates communication effectiveness through chart type appropriateness, encoding clarity, and insight salience. \VJudge scores each chart on a 0–5 scale with a detailed rubric.

\textit{(c) Aesthetics (AE).}
This dimension assesses visual appeal through color harmony, layout quality, and absence of visual noise, anchored by calibration examples of exemplary and poor designs. \VJudge scores each chart on a 0–5 scale.

\textit{(d) Technical Correctness (TC).}
This dimension applies rule-based checks for missing chart elements, scale errors (including the critical zero-baseline check), label readability issues, and color accessibility problems. This provides objective, reproducible diagnostics. \VJudge scores each chart on a 0–5 scale. The final VisScore is the average of \text{DF}, \text{EX}, \text{AE} and \text{TC}.

\textbf{(3) Overall Score.}
The overall \bench score is a weighted combination of \TJudge and \VJudge scores, with a default weighting of 0.6 for \TJudge and 0.4 for \VJudge, reflecting the typical priority in business reporting where textual analysis carries the primary analytical burden while visualizations provide essential support. \textbf{All report dimensions and the overall score are normalized to [0,100] for final score}. The implementation details of dual-judge are provided in Appendix~\ref{app:judge}.

\section{Experiments}
\label{sec:experiments}


\subsection{Experimental Setup}
\label{sec:exp_setup}
\textbf{Evaluation Protocol.}
For text-only models, only the textual component is generated, and the visual score is set to zero. For multimodal models, chart generation can be triggered via code execution or native multimodal output, depending on model capabilities. All models use the same prompt template and are allowed a maximum of 4096 tokens for the textual report and up to 8 visualizations. The reference reports and keypoints serve as ground truth for evaluation. The judge LLMs (GPT-4o and GPT-4o with vision) are used for \TJudge and \VJudge scoring at temperature 0.2 to maximize reproducibility. For Metrics, we report \TJudge scores, \VJudge scores, and the overall score. Each report is
evaluated three times, and the average scores are reported.

\textbf{Baseline Models.}
We evaluate 20 state-of-the-art models spanning four categories: (i) text-only LLMs (GPT-4o-text~\cite{achiam2023gpt}, Claude-3.5-Sonnet~\cite{anthropic2024claudesonnet35}, DeepSeek-V3~\cite{guo2025deepseek}, Qwen3-32B, Llama-3.3-70B, GPT-5.5~\cite{achiam2023gpt}), (ii) vision-language models with native chart generation (GPT-4o~\cite{achiam2023gpt}, GPT-4.1, Gemini-2.5-Pro, Gemini-3-Flash, Claude-4.5-Sonnet~\cite{anthropic2025claudesonnet45}, Qwen3-VL-235B~\cite{qwen3-vl-235b}, Gemini 3.5-Flash, Claude Opus 4.7~\cite{anthropic-claude-opus-4.7}), (iii) code-augmented multimodal systems that generate charts via Python execution (DeepSeek-R1 with code interpreter~\cite{guo2025deepseek}, Qwen3-Coder, TableGPT2~\cite{su2024tablegpt2}), and (iv) specialized deep research agents (Gemini Deep Research~\cite{google-gemini-deep-research}, ChatGPT Deep Research, Perplexity Sonar Deep Research). All models are accessed via official APIs or open-source inference between November 2025 and January 2026.


\subsection{Main Results}
\label{sec:main_results}

Table~\ref{tab:main_results} presents the overall and dimension-level performance of all evaluated models across the all 386 tasks. We highlight several key findings.

\textbf{Finding 1: Text-only LLMs are fundamentally insufficient for multimodal business reporting.} Without visualization capabilities, these models score zero on \VJudge, capping their overall \bench score below 24. Even in textual dimensions, they lag behind multimodal counterparts in GIV (up to 10 points lower), likely because visual thinking aids in data comprehension and fact checking.

\textbf{Finding 2: Code-augmented multimodal systems lead, but the gap is modest.} DeepSeek-R1 with code interpreter achieves the highest overall score (68.9), outperforming the strongest vision-language model (Gemini-2.5-Pro at 66.8) by 2.1 points. The advantage is most pronounced in Data Fidelity (85.1 vs 82.4) and Expressiveness (78.3 vs 76.1), attributed to the precise control that code execution provides over chart rendering. However, even this best model achieves only 68.9, far from human-level performance (human baseline: 92.3, see section~\ref{sec:human}, which is computed from independently human-written multimodal reports, not from the curated reference reports), indicating substantial room for improvement.

\textbf{Finding 3: All models struggle with insight novelty.} The Novelty Analysis (NA) scores are markedly lower than SC and GIV across the board. DeepSeek-R1, the best performer, reaches only 38.7 on NA, revealing that current models predominantly rely on paraphrasing existing data patterns and rarely produce genuinely non-trivial, actionable insights. This validates our inclusion of NA as a critical dimension absent from prior benchmarks.

\textbf{Finding 4: Visualization quality remains a bottleneck.} Even the top \VJudge scores hover around 80, with Data Fidelity being the strongest dimension (85.1 for DeepSeek-R1) and Aesthetics consistently weaker (max 76.0). Technical correctness issues (e.g., missing axis labels, unreadable legends) are frequent, especially among multimodal models without explicit code execution.

\begin{table*}[t!]
\centering
\caption{Overall performance on \bench. We report the main scores and the Overall score (Overall). The best results in each column are \textbf{bolded}, the second-best are \underline{underlined}.}
\label{tab:main_results}
\resizebox{\textwidth}{!}{%
\begin{tabular}{lccccccccccc}
\toprule
\textbf{Model} & \textbf{SC} & \textbf{GIV} & \textbf{NA} & \textbf{CMC} & \textbf{T-Score} & \textbf{DF} & \textbf{EX} & \textbf{AE} & \textbf{TC} & \textbf{V-Score} & \textbf{Overall} \\
\midrule
\multicolumn{11}{c}{\textit{Text-only LLMs}} \\
GPT-4o (text)       & 72.3 & 48.5 & 28.1 & 0.0  & 37.2 & --- & --- & --- & --- & 0.0  & 22.3 \\
Claude-3.5-Sonnet   & 70.8 & 45.2 & 25.6 & 0.0  & 35.4 & --- & --- & --- & --- & 0.0  & 21.2 \\
DeepSeek-V3         & 68.4 & 46.1 & 27.3 & 0.0  & 35.5 & --- & --- & --- & --- & 0.0  & 21.3 \\
Qwen3-32B           & 67.2 & 43.8 & 24.5 & 0.0  & 33.9 & --- & --- & --- & --- & 0.0  & 20.3 \\
Llama-3.3-70B       & 65.9 & 41.2 & 22.7 & 0.0  & 32.5 & --- & --- & --- & --- & 0.0  & 19.5 \\
GPT-5.5             & 72.9 & 51.2 & 32.7 & 0.0  & 39.2 & --- & --- & --- & --- & 0.0  & 23.5 \\
\midrule
\multicolumn{11}{c}{\textit{Vision-Language Models}} \\
GPT-4o              & 75.4 & 55.8 & 34.2 & 64.1 & 57.4 & 78.2 & 72.5 & 68.3 & 82.1 & 74.8 & 64.3 \\
GPT-4.1             & 74.8 & 54.6 & 32.8 & 62.5 & 56.1 & 76.5 & 70.8 & 66.9 & 80.4 & 73.2 & 63.0 \\
Gemini-2.5-Pro      & 76.1 & 57.3 & 35.5 & 67.2 & 59.0 & 82.4 & 76.1 & 73.0 & 84.6 & 78.4 & 66.8 \\
Gemini-3-Flash      & 73.5 & 53.1 & 32.1 & 63.5 & 55.6 & 79.8 & 73.2 & 70.5 & 82.8 & 76.1 & 63.8 \\
Gemini-3.5-Flash    & 76.2 & 54.6 & 33.3 & 65.3 & 57.3 & 80.6 & 72.9 & 71.9 & 82.7 & 76.6 & 65.0 \\
Claude-4.5-Sonnet   & 74.2 & 54.8 & 33.6 & 64.5 & 56.7 & 80.3 & 74.0 & 72.1 & 83.3 & 76.8 & 64.8 \\
Claude Opus 4.7     & 68.2 & 53.1 & 34.2 & 63.1 & 54.6 & 78.8 & 72.8 & 72.8 & 82.6 & 77.1 & 63.6 \\
Qwen3-VL-235B       & 71.6 & 51.9 & 30.4 & 61.0 & 53.7 & 77.9 & 71.5 & 67.8 & 79.2 & 73.0 & 61.4 \\
\midrule
\multicolumn{11}{c}{\textit{Code-augmented Multimodal Systems}} \\
DeepSeek-R1         & \textbf{78.2} & \textbf{59.4} & \textbf{38.7} & \textbf{69.1} & \textbf{61.35} & \textbf{85.1} & \textbf{78.3} & \underline{74.2} & \textbf{86.5} & \textbf{80.2} & \textbf{68.9} \\
Qwen3-Coder-32B     & 72.9 & 52.4 & 31.5 & 63.1 & 55.0 & 80.5 & 74.1 & 71.3 & 83.0 & 76.5 & 63.6 \\
TableGPT2-7B        & 65.3 & 44.7 & 23.2 & 52.3 & 46.3 & 68.9 & 62.5 & 58.4 & 71.2 & 64.0 & 53.4 \\
\midrule
\multicolumn{11}{c}{\textit{Deep Research Agents}} \\
Gemini Deep Research    & \underline{77.5} & \underline{58.1} & \underline{37.2} & \underline{68.1} & \underline{60.2} & \underline{83.1} & \underline{77.5} & \textbf{76.0} & \underline{85.8} & \underline{79.6} & \underline{67.9} \\
ChatGPT Deep Research   & 75.2 & 55.9 & 35.0 & 65.5 & 57.9 & 80.9 & 74.8 & 73.5 & 83.9 & 77.3 & 65.7 \\
Perplexity Sonar DR     & 73.8 & 53.5 & 33.8 & 63.5 & 56.2 & 79.0 & 72.9 & 70.8 & 82.5 & 75.5 & 63.9 \\
\bottomrule
\end{tabular}%
}
\end{table*}

\begin{figure*}[t]
\centering
\includegraphics[width=1\linewidth]{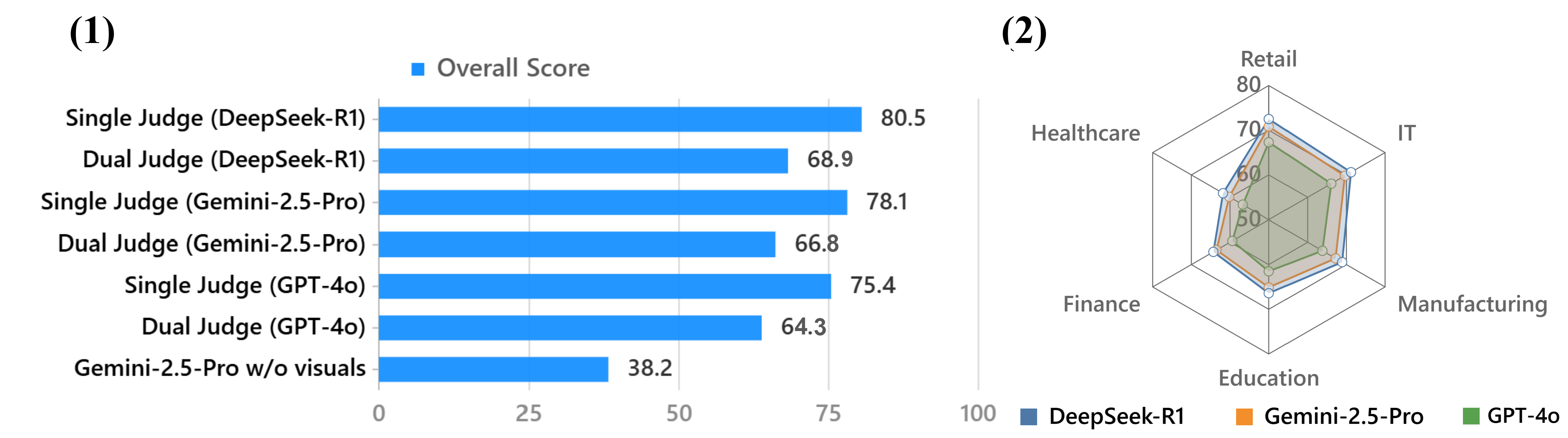}
\caption{(1) shows the overall scores of three representative models under Single Judge vs. Dual-Judg. Dual-Judge's scores are consistently lower, revealing a score inflation of roughly 8–12 points in the single‑judge setting. The chart also illustrates the severe performance drop when visual generation is disabled (w/o visuals). (2) is a radar chart of domain-level \bench scores for the top three models. DeepSeek-R1 maintains a lead across all domains, but all models exhibit a similar ``hard domain'' pattern: Finance and Healthcare are consistently more difficult than Retail and IT Operations.}
\label{fig:charts}
\end{figure*}

\subsection{Ablation Study: Modality and Evaluation Configurations}
\label{sec:ablation}


\textbf{Impact of Visual Generation.}
To quantify the value of multimodal generation, we compare the performance of the same base model (Gemini-2.5-Pro) when generating reports with vs.\ without charts. As shown in Figure~\ref{fig:charts}(1), removing visualizations (text-only setting) drops the overall score from 66.8 to 38.2, but interestingly, the \TJudge sub-scores also degrade (GIV from 57.3 to 49.1, NA from 35.5 to 29.3). This suggests that the process of creating charts may force the model to better verify its data, indirectly improving textual factuality. Conversely, a purely visual report without text scores only 46.5 overall, highlighting the complementarity of modalities.

\textbf{Impact of Dual-Judge Design.}
To validate the need for a separate \VJudge, we experiment with a \emph{unified judge} (single GPT-4o prompt that scores both text and visuals together). The result is shown in Figure~\ref{fig:charts} (1). The unified judge consistently inflates overall scores by 8--12 points compared to our dual-judge framework, and its correlation with human expert judgments drops from $r=0.89$ to $r=0.72$. This confirms that splitting the evaluation into specialized judges improves accuracy and reduces the halo effect (where good text masks poor charts, and vice versa), consistent with the concerns about single-judge fragility.

\subsection{Performance Analysis}
\textbf{Domain-Specific Analysis.}
\label{sec:domain_analysis}
Figure~\ref{fig:charts} (2) shows the per-domain performance of the three best models. All three models perform best in \emph{Retail} and \emph{IT Operations}, where tables tend to be simpler and instructions are more straightforward. Performance drops sharply in \emph{Healthcare} (complex temporal patient data, heavy numerical content) and \emph{Finance} (multi-table consolidation, dense accounting metrics). For example, DeepSeek-R1 achieves 72.5 in Retail but only 61.8 in Healthcare, a gap of over 10 points. The main degradation factors are hallucinated numerical values in healthcare (adverse events misattributed) and structural confusion in multi-table financial reports (balance sheet and income statement items mixed). This domain sensitivity underscores the need for domain-specific adaptation strategies.

\textbf{Human Evaluation and Consistency.}
\label{sec:human}
We conduct a human evaluation to assess the alignment of our automated metrics with expert judgment. Six professional business analysts evaluate a stratified sample of 100 generated reports (covering all domains and top models). They score each report on the same dimensions as the \TJudge and \VJudge, following the rubrics described in Section~\ref{sec:evaluation}. Inter-annotator agreement reaches Fleiss' $\kappa = 0.83$, indicating high consistency. As shown in Appendix Table~\ref{tab:human_corr}, the \bench automated scores achieve a Pearson correlation of $0.908$ with human overall ratings, and dimension-level correlations range from $0.82$ to $0.94$. Notably, the automated system tends to be \emph{stricter} than humans (average score difference $\sim3.5$ points), likely because automated numerical verification catches subtle hallucinations that humans may overlook. This conservative bias is preferable for a benchmark intended to drive improvement.

\textbf{Error Analysis}
\label{subsec:error}
We manually analyze 200 randomly sampled reports to categorize failure modes. The error taxonomy is extended for multimodal settings. Appendix Table~\ref{tab:error_types} shows the distribution of primary error types in generated report.
Numerical hallucination remains the most pervasive issue, occurring in 38.5\% of reports. Hallucinated values often appear plausible but are off by an order of magnitude or swapped between columns. Trivial insights (31\%) reflect the model's tendency to state obvious facts (e.g., ``sales increased'') without explaining why or providing context. Chart-data mismatch (22\%) includes cases where a line chart claims a growth rate of 20\%, but the underlying table shows 5\%. Multi-table confusion (14\%) is especially prevalent in finance reports with multiple interconnected statements. Truncation occurs in 12.5\% of reports involving extremely large tables, where models lose context after the maximum token limit, leading to incomplete or incorrect analysis.

\section{Conclusion}
We introduced \bench, the first unified multimodal benchmark for table-to-report generation with a comprehensive dual-judge evaluation framework. It contains 386 expert-curated tasks across six business domains, together with reference reports and verifiable key insight points. Reference reports are constructed through an MCTS-based generation pipeline followed by multi-stage human curation and verification. Our work is motivated by and directly addresses the key limitations identified across prior research. Extensive experiments on 20 models reveal that multimodal business report generation remains an open challenge. The best model achieves 68.9 overall score versus the human baseline of 92.3, with insight novelty and visualization quality as the major bottlenecks.Our ablation further confirms that a single unified judge inflates scores by 8–12 points, whereas the dual-judge design reduces cross-modality compensation and aligns better with human expert judgment.
We hope \bench provides a rigorous testbed for evaluating and advancing reliable multimodal report generation. Future work includes improving numerical faithfulness, novelty-oriented insight generation, chart fidelity, multi-table reasoning, domain adaptation, and evaluation protocols robust to reward hacking.

\section{AI use statement}
In this work, we used generative AI tools for the following tasks with required disclosure: generating candidate analytical queries from table metadata; generating candidate reference reports through the MCTS pipeline; serving as LLM judges for T-Judge, V-Judge, and cross-modal consistency evaluation; and generating SQL/code for numerical verification of claims and keypoints. We have not used generative AI tools for the following required-disclosure tasks: formulating or proving mathematical claims, proposing or refining hypotheses, designing the core research methodology, conducting qualitative or thematic data analysis, translating data, cleaning or reformatting datasets, or interpreting results. Other required-disclosure tasks not listed above were not performed with generative AI. Additionally, we used generative AI tools for recommended-disclosure tasks, including editing the manuscript for readability, formatting references, suggesting paper structure and title, and assisting with code for data processing and evaluation. All AI-assisted work was reviewed and verified by the authors. In particular, LLM-generated queries, reference candidates, keypoints, charts, and evaluation outputs were checked against source tables and/or expert judgment; the final human-curated reference reports and keypoint annotations were produced by human annotators. We take responsibility for the final content of this work, including text, claims, and artifacts produced with the aid of generative AI.

\bibliographystyle{plainnat}
\bibliography{refs/custom}

\begin{thebibliography}{37}
\providecommand{\natexlab}[1]{#1}
\providecommand{\url}[1]{\texttt{#1}}
\expandafter\ifx\csname urlstyle\endcsname\relax
  \providecommand{\doi}[1]{doi: #1}\else
  \providecommand{\doi}{doi: \begingroup \urlstyle{rm}\Url}\fi

\bibitem[ant(2025)]{anthropic-claude-opus-4.7}
Claude opus 4.7.
\newblock Anthropic Official Announcement, 2025.
\newblock URL \url{https://www.anthropic.com/news/claude-opus-4-7}.
\newblock Accessed: 2026-07-27.

\bibitem[goo(2025)]{google-gemini-deep-research}
Gemini deep research.
\newblock Google Gemini, 2025.
\newblock URL \url{https://gemini.google.com/}.
\newblock Accessed: 2026-07-27.

\bibitem[qwe(2025)]{qwen3-vl-235b}
Qwen3-vl-235b.
\newblock Qwen AI, 2025.
\newblock URL \url{https://qwen.ai/}.
\newblock Accessed: 2026-07-27.

\bibitem[Achiam et~al.(2023)Achiam, Adler, Agarwal, Ahmad, Akkaya, Aleman, Almeida, Altenschmidt, Altman, Anadkat, et~al.]{achiam2023gpt}
Josh Achiam, Steven Adler, Sandhini Agarwal, Lama Ahmad, Ilge Akkaya, Florencia~Leoni Aleman, Diogo Almeida, Janko Altenschmidt, Sam Altman, Shyamal Anadkat, et~al.
\newblock Gpt-4 technical report.
\newblock \emph{arXiv preprint arXiv:2303.08774}, 2023.

\bibitem[{Anthropic}(2024)]{anthropic2024claudesonnet35}
{Anthropic}.
\newblock Claude 3.5.
\newblock Anthropic Blog, 2024.
\newblock URL \url{https://www.anthropic.com/news/claude-3-5-sonnet}.
\newblock Introducing Claude Sonnet 3.5.

\bibitem[{Anthropic}(2025)]{anthropic2025claudesonnet45}
{Anthropic}.
\newblock Claude 4.5.
\newblock Anthropic Blog, 2025.
\newblock URL \url{https://www.anthropic.com/news/claude-sonnet-4-5}.
\newblock Introducing Claude Sonnet 4.5, released 29 Sep 2025.

\bibitem[Chen et~al.(2024)Chen, Zhang, Xu, Ren, and Yang]{chen2024viseval}
Nan Chen, Yuge Zhang, Jiahang Xu, Kan Ren, and Yuqing Yang.
\newblock Viseval: A benchmark for data visualization in the era of large language models.
\newblock \emph{IEEE Transactions on Visualization and Computer Graphics}, 31\penalty0 (1):\penalty0 1301--1311, 2024.

\bibitem[Chen et~al.(2025{\natexlab{a}})Chen, Hu, Zou, Wu, Wang, Hooi, and He]{chen2025judgelrm}
Nuo Chen, Zhiyuan Hu, Qingyun Zou, Jiaying Wu, Qian Wang, Bryan Hooi, and Bingsheng He.
\newblock Judgelrm: Large reasoning models as a judge.
\newblock \emph{arXiv preprint arXiv:2504.00050}, 2025{\natexlab{a}}.

\bibitem[Chen et~al.(2020)Chen, Wang, Chen, Zhang, Wang, Li, Zhou, and Wang]{Chen2020TabFact}
Wenhu Chen, Hongmin Wang, Jianshu Chen, Yunkai Zhang, Hong Wang, Shiyang Li, Xiyou Zhou, and William~Yang Wang.
\newblock Tabfact: A large-scale dataset for table-based fact verification.
\newblock In \emph{International Conference on Learning Representations}, 2020.
\newblock URL \url{https://openreview.net/forum?id=rkeJRhNYDH}.

\bibitem[Chen et~al.(2025{\natexlab{b}})Chen, Li, Wang, Jin, Qian, Wang, Wang, Zhang, Zhang, Zhang, et~al.]{chen2025rm}
Xiusi Chen, Gaotang Li, Ziqi Wang, Bowen Jin, Cheng Qian, Yu~Wang, Hongru Wang, Yu~Zhang, Denghui Zhang, Tong Zhang, et~al.
\newblock Rm-r1: Reward modeling as reasoning.
\newblock \emph{arXiv preprint arXiv:2505.02387}, 2025{\natexlab{b}}.

\bibitem[Cheng et~al.(2022)Cheng, Dong, Wang, Jia, Guo, Gao, Han, Lou, and Zhang]{cheng2022hitab}
Zhoujun Cheng, Haoyu Dong, Zhiruo Wang, Ran Jia, Jiaqi Guo, Yan Gao, Shi Han, Jian-Guang Lou, and Dongmei Zhang.
\newblock Hitab: A hierarchical table dataset for question answering and natural language generation.
\newblock In \emph{Proceedings of the 60th Annual Meeting of the Association for Computational Linguistics (Volume 1: Long Papers)}, pages 1094--1110, 2022.

\bibitem[Coulom(2007)]{mcts}
R{\'e}mi Coulom.
\newblock Efficient selectivity and backup operators in monte-carlo tree search.
\newblock In H.~Jaap van~den Herik, Paolo Ciancarini, and H.~H. L. M.~(Jeroen) Donkers, editors, \emph{Computers and Games}, pages 72--83, Berlin, Heidelberg, 2007. Springer Berlin Heidelberg.
\newblock ISBN 978-3-540-75538-8.

\bibitem[Guo et~al.(2025)Guo, Yang, Zhang, Song, Wang, Zhu, Xu, Zhang, Ma, Bi, et~al.]{guo2025deepseek}
Daya Guo, Dejian Yang, Haowei Zhang, Junxiao Song, Peiyi Wang, Qihao Zhu, Runxin Xu, Ruoyu Zhang, Shirong Ma, Xiao Bi, et~al.
\newblock Deepseek-r1 incentivizes reasoning in llms through reinforcement learning.
\newblock \emph{Nature}, 645\penalty0 (8081):\penalty0 633--638, 2025.

\bibitem[Huang et~al.(2026)Huang, Zhong, Wan, Zhou, Alam, Wang, Li, Dou, Zhu, Xiong, Tao, Xu, Dimitriadis, Zhang, and Zhang]{huang2026mmdeepresearch}
Peizhou Huang, Zixuan Zhong, Zhongwei Wan, Donghao Zhou, Samiul Alam, Xin Wang, Zexin Li, Zhihao Dou, Li~Zhu, Jing Xiong, Chaofan Tao, Yan Xu, Dimitrios Dimitriadis, Tuo Zhang, and Mi~Zhang.
\newblock Mmdeepresearch-bench: A benchmark for multimodal deep research agents, 2026.
\newblock URL \url{https://arxiv.org/abs/2601.12346}.

\bibitem[Kocsis and Szepesv{\'a}ri(2006)]{uct}
Levente Kocsis and Csaba Szepesv{\'a}ri.
\newblock Bandit based monte-carlo planning.
\newblock In Johannes F{\"u}rnkranz, Tobias Scheffer, and Myra Spiliopoulou, editors, \emph{Machine Learning: ECML 2006}, pages 282--293, Berlin, Heidelberg, 2006. Springer Berlin Heidelberg.
\newblock ISBN 978-3-540-46056-5.

\bibitem[Lin(2025{\natexlab{a}})]{lin2025Simplifying}
Teng Lin.
\newblock Simplifying data integration: Slm-driven systems for unified semantic queries across heterogeneous databases.
\newblock In \emph{2025 IEEE 41st International Conference on Data Engineering (ICDE)}, pages 4690--4693, May 2025{\natexlab{a}}.
\newblock \doi{10.1109/ICDE65448.2025.00378}.

\bibitem[Lin(2025{\natexlab{b}})]{lin2025lightkggsimpleefficientknowledge}
Teng Lin.
\newblock Lightkgg: Simple and efficient knowledge graph generation from textual data, 2025{\natexlab{b}}.
\newblock URL \url{https://arxiv.org/abs/2510.23341}.

\bibitem[Lin(2025{\natexlab{c}})]{lin2025structured}
Teng Lin.
\newblock Structured retrieval-augmented generation for multi-entity question answering over heterogeneous sources.
\newblock In \emph{2025 IEEE 41st International Conference on Data Engineering Workshops (ICDEW)}, pages 253--258, 2025{\natexlab{c}}.
\newblock \doi{10.1109/ICDEW67478.2025.00036}.

\bibitem[Lin et~al.(2025{\natexlab{a}})Lin, Luo, Zhang, Zhang, Liu, Wu, and Tang]{lin-etal-2025-mebench}
Teng Lin, Yuyu Luo, Honglin Zhang, Jicheng Zhang, Chunlin Liu, Kaishun Wu, and Nan Tang.
\newblock {MEB}ench: Benchmarking large language models for cross-document multi-entity question answering.
\newblock In Christos Christodoulopoulos, Tanmoy Chakraborty, Carolyn Rose, and Violet Peng, editors, \emph{Proceedings of the 2025 Conference on Empirical Methods in Natural Language Processing}, pages 1481--1494, Suzhou, China, November 2025{\natexlab{a}}. Association for Computational Linguistics.
\newblock ISBN 979-8-89176-332-6.
\newblock \doi{10.18653/v1/2025.emnlp-main.77}.
\newblock URL \url{https://aclanthology.org/2025.emnlp-main.77/}.

\bibitem[Lin et~al.(2025{\natexlab{b}})Lin, Zhu, Luo, and Tang]{lin2025srag}
Teng Lin, Yizhang Zhu, Yuyu Luo, and Nan Tang.
\newblock Srag: Structured retrieval-augmented generation for multi-entity question answering over wikipedia graph.
\newblock \emph{CoRR}, abs/2503.01346, March 2025{\natexlab{b}}.
\newblock URL \url{https://doi.org/10.48550/arXiv.2503.01346}.

\bibitem[Lin et~al.(2026{\natexlab{a}})Lin, Liu, and Tang]{lin2026cominsight}
Teng Lin, Xinyu Liu, and Nan Tang.
\newblock Structured composition of verifiable atomic insights for table-to-report generation, 2026{\natexlab{a}}.
\newblock URL \url{https://arxiv.org/abs/2610.03525}.

\bibitem[Lin et~al.(2026{\natexlab{b}})Lin, Luo, and Tang]{lin2026annoretrieveefficientstructuredretrieval}
Teng Lin, Yuyu Luo, and Nan Tang.
\newblock Annoretrieve: Efficient structured retrieval for unstructured document analysis, 2026{\natexlab{b}}.
\newblock URL \url{https://arxiv.org/abs/2604.02690}.

\bibitem[Lin et~al.(2026{\natexlab{c}})Lin, Luo, and Tang]{lin2026structurethenquery}
Teng Lin, Yuyu Luo, and Nan Tang.
\newblock Structure then query: Enabling precise analytical queries over unstructured documents, 2026{\natexlab{c}}.
\newblock URL \url{https://arxiv.org/abs/2608.13384}.

\bibitem[Lin et~al.(2026{\natexlab{d}})Lin, Zhang, Luo, and Tang]{lin2026montecarlotreesearch}
Teng Lin, Zhiyang Zhang, Yuyu Luo, and Nan Tang.
\newblock Monte carlo tree search for table-to-multimodal report generation, 2026{\natexlab{d}}.
\newblock URL \url{https://arxiv.org/abs/2608.04071}.

\bibitem[Lin et~al.(2026{\natexlab{e}})Lin, Zhu, Zhang, Luo, and Tang]{lin2026docsageinformationstructuringagent}
Teng Lin, Yizhang Zhu, Zhengxuan Zhang, Yuyu Luo, and Nan Tang.
\newblock Docsage: An information structuring agent for multi-doc multi-entity question answering, 2026{\natexlab{e}}.
\newblock URL \url{https://arxiv.org/abs/2603.11798}.

\bibitem[Liu et~al.(2026{\natexlab{a}})Liu, Yu, Orini, Du, and He]{liu2026huntinstead}
Wei Liu, Peijie Yu, Michele Orini, Yali Du, and Yulan He.
\newblock Hunt instead of wait: Evaluating deep data research on large language models, 2026{\natexlab{a}}.
\newblock URL \url{https://arxiv.org/abs/2602.02039}.

\bibitem[Liu et~al.(2026{\natexlab{b}})Liu, Yu, Su, Wang, Wang, Jiang, Liu, Cohan, Tian, and Chen]{liu2026examining}
Yixin Liu, Yue Yu, DiJia Su, Sid Wang, Xuewei Wang, Song Jiang, Bo~Liu, Arman Cohan, Yuandong Tian, and Zhengxing Chen.
\newblock Examining reasoning llms-as-judges in non-verifiable llm post-training, 2026{\natexlab{b}}.
\newblock URL \url{https://arxiv.org/abs/2603.12246}.

\bibitem[Nan et~al.(2022)Nan, Hsieh, Mao, Lin, Verma, Zhang, Kry{\'s}ci{\'n}ski, Schoelkopf, Kong, Tang, et~al.]{nan2022fetaqa}
Linyong Nan, Chiachun Hsieh, Ziming Mao, Xi~Victoria Lin, Neha Verma, Rui Zhang, Wojciech Kry{\'s}ci{\'n}ski, Hailey Schoelkopf, Riley Kong, Xiangru Tang, et~al.
\newblock Fetaqa: Free-form table question answering.
\newblock \emph{Transactions of the Association for Computational Linguistics}, 10:\penalty0 35--49, 2022.

\bibitem[Pan et~al.(2025)Pan, Fu, Wang, Lu, Pan, Qian, Chen, Wang, Zhou, Zheng, et~al.]{pan2025vis}
Bo~Pan, Yixiao Fu, Ke~Wang, Junyu Lu, Lunke Pan, Ziyang Qian, Yuhan Chen, Guoliang Wang, Yitao Zhou, Li~Zheng, et~al.
\newblock Vis-shepherd: Constructing critic for llm-based data visualization generation.
\newblock \emph{arXiv preprint arXiv:2506.13326}, 2025.

\bibitem[Parikh et~al.(2020)Parikh, Wang, Gehrmann, Faruqui, Dhingra, Yang, and Das]{parikh2020totto}
Ankur Parikh, Xuezhi Wang, Sebastian Gehrmann, Manaal Faruqui, Bhuwan Dhingra, Diyi Yang, and Dipanjan Das.
\newblock Totto: A controlled table-to-text generation dataset.
\newblock In \emph{Proceedings of the 2020 Conference on Empirical Methods in Natural Language Processing (EMNLP)}, pages 1173--1186, 2020.

\bibitem[Pasupat and Liang(2015)]{pasupat2015compositional}
Panupong Pasupat and Percy Liang.
\newblock Compositional semantic parsing on semi-structured tables.
\newblock In \emph{Proceedings of the 53rd Annual Meeting of the Association for Computational Linguistics and the 7th International Joint Conference on Natural Language Processing (Volume 1: Long Papers)}, pages 1470--1480, 2015.

\bibitem[Su et~al.(2024)Su, Wang, Ye, Zhou, Zhang, Zhu, Wang, Xu, Chen, Li, Lan, Tian, Yuan, Zhao, Zhou, Shou, Zha, Long, Li, Wu, Zhang, Huang, Yang, Zhang, Ye, Zhu, Hu, Gu, Sun, Li, Yang, and Xiao]{su2024tablegpt2}
Aofeng Su, Aowen Wang, Chao Ye, Chen Zhou, Ga~Zhang, Guangcheng Zhu, Haobo Wang, Haokai Xu, Hao Chen, Haoze Li, Haoxuan Lan, Jiaming Tian, Jing Yuan, Junbo Zhao, Junlin Zhou, Kaizhe Shou, Liangyu Zha, Lin Long, Liyao Li, Pengzuo Wu, Qi~Zhang, Qingyi Huang, Saisai Yang, Tao Zhang, Wentao Ye, Wufang Zhu, Xiaomeng Hu, Xijun Gu, Xinjie Sun, Xiang Li, Yuhang Yang, and Zhiqing Xiao.
\newblock Tablegpt2: A large multimodal model with tabular data integration, 2024.
\newblock URL \url{https://arxiv.org/abs/2411.02059}.

\bibitem[Wu et~al.(2025)Wu, Yang, Chai, Zhang, Liu, Du, Liang, Shu, Cheng, Sun, et~al.]{wu2025tablebench}
Xianjie Wu, Jian Yang, Linzheng Chai, Ge~Zhang, Jiaheng Liu, Xeron Du, Di~Liang, Daixin Shu, Xianfu Cheng, Tianzhen Sun, et~al.
\newblock Tablebench: A comprehensive and complex benchmark for table question answering.
\newblock In \emph{Proceedings of the AAAI Conference on Artificial Intelligence}, volume~39, pages 25497--25506, 2025.

\bibitem[Xie et~al.(2026)Xie, Zhang, Wu, Lu, Zhang, Yu, Wang, Hong, Liu, Wu, and Luo]{xie2025visjudge}
Yupeng Xie, Zhiyang Zhang, Yifan Wu, Sirong Lu, Jiayi Zhang, Zhaoyang Yu, Jinlin Wang, Sirui Hong, Bang Liu, Chenglin Wu, and Yuyu Luo.
\newblock Visjudge-bench: Aesthetics and quality assessment of visualizations.
\newblock In \emph{The Fourteenth International Conference on Learning Representations}, 2026.
\newblock URL \url{https://openreview.net/forum?id=lG1HWWdEbN}.

\bibitem[Yu et~al.(2018)Yu, Zhang, Yang, Yasunaga, Wang, Li, Ma, Li, Yao, Roman, et~al.]{yu2018spider}
Tao Yu, Rui Zhang, Kai Yang, Michihiro Yasunaga, Dongxu Wang, Zifan Li, James Ma, Irene Li, Qingning Yao, Shanelle Roman, et~al.
\newblock Spider: A large-scale human-labeled dataset for complex and cross-domain semantic parsing and text-to-sql task.
\newblock In \emph{Proceedings of the 2018 conference on empirical methods in natural language processing}, pages 3911--3921, 2018.

\bibitem[Zhang et~al.(2025)Zhang, Pan, Xiong, Wei, Zhao, Li, Peng, Gu, Yang, Chang, Wu, Zhong, Song, and Li]{zhang2025t2rbench}
Jie Zhang, Changzai Pan, Sishi Xiong, Kaiwen Wei, Yu~Zhao, Xiangyu Li, Jiaxin Peng, Xiaoyan Gu, Jian Yang, Wenhan Chang, Zhenhe Wu, Jiang Zhong, Shuangyong Song, and Xuelong Li.
\newblock {T}2{R}-{BENCH}: A benchmark for real world table-to-report task.
\newblock In Christos Christodoulopoulos, Tanmoy Chakraborty, Carolyn Rose, and Violet Peng, editors, \emph{Proceedings of the 2025 Conference on Empirical Methods in Natural Language Processing}, pages 22427--22451, Suzhou, China, November 2025. Association for Computational Linguistics.
\newblock ISBN 979-8-89176-332-6.
\newblock \doi{10.18653/v1/2025.emnlp-main.1141}.
\newblock URL \url{https://aclanthology.org/2025.emnlp-main.1141/}.

\bibitem[Zhong et~al.(2017)Zhong, Xiong, and Socher]{zhong2017seq2sql}
Victor Zhong, Caiming Xiong, and Richard Socher.
\newblock Seq2sql: Generating structured queries from natural language using reinforcement learning.
\newblock \emph{arXiv preprint arXiv:1709.00103}, 2017.

\end{thebibliography}
\appendix
\section{Appendix}
\label{sec:appendix}

\subsection{Related work}
\textbf{Autonomous Data Analysis and Deep Research.}
A parallel line of work concerns agentic data analysis and deep research, where models must autonomously decide what to investigate and produce comprehensive findings. DDR-Bench~\cite{liu2026huntinstead} formalizes \textit{investigatory intelligence} as the ability of LLMs to explore structured databases without explicit questions, and demonstrates that current agents suffer from limited exploration breadth, premature termination, and weak insight synthesis, even when scaffolded with tool use. MMDR-Bench~\cite{huang2026mmdeepresearch} extends this to multimodal deep research agents that must coordinate web browsing, evidence gathering, and report writing with visual grounding. Both benchmarks highlight critical bottlenecks: factual hallucination, insufficient coverage of key evidence, and a trade-off between writing fluency and evidential faithfulness. Our \bench fills this gap by uniting the structured-table grounding of T2R-Bench with the exploratory depth of DDR-Bench and the multimodal verification rigor of MMDR-Bench, and by introducing a standardized reference-report structure that enables fine-grained, reproducible evaluation across six real-world business domains.

\subsection{MCTS}
\label{subsec:MCTS}
We formulate report generation as a sequential decision-making process over an action space $\mathcal{A}$ consisting of two stages with six action types:

\begin{itemize}
    \item \textbf{Analysis Stage Actions:}
    \begin{itemize}
        \item $A_1$: \textbf{Chapter Plan} --- Propose a chapter-level outline for the report based on the query and table data.
        \item $A_2$: \textbf{Visualization Task} --- Identify specific chart-worthy analytical points and specify chart types, data filters, and encodings.
        \item $A_3$: \textbf{Chart Generation} --- Generate one or more visualizations (bar charts, line plots, pie charts, etc.) via code execution or native multimodal output.
        \item $A_4$: \textbf{Chart Modification} --- Refine previously generated charts (e.g., adjust scales, fix labeling, improve color schemes) based on self-critique.
        \item $A_5$: \textbf{Organize Insight} --- Draft a textual insight paragraph grounded in the table data and linked to the supporting chart(s).
        \item $A_6$: \textbf{Chapter Summary} --- Synthesize all insights within a chapter into a cohesive section.
    \end{itemize}
    \item \textbf{Refinement Stage Actions:}
    \begin{itemize}
        \item $R_1$: \textbf{Title Optimization} --- Refine the report title and section headings for clarity and impact.
        \item $R_2$: \textbf{Narrative Flow} --- Improve transitions between paragraphs and chapters for coherent reading.
        \item $R_3$: \textbf{Transition Design} --- Add bridging sentences and forward-looking statements between major sections.
    \end{itemize}
\end{itemize}

The MCTS procedure follows four phases:

\begin{enumerate}
    \item \textbf{Selection:} Starting from the root, recursively select child nodes with the highest UCT value until reaching an expandable node.
    \item \textbf{Expansion:} Create one or more child nodes by applying an LLM-action from $\mathcal{A}$ to the selected node's report state.
    \item \textbf{Simulation:} From the newly expanded node, complete the report to a terminal state by sequentially applying the remaining actions in a predefined order (analysis stage first, then refinement stage). To encourage diversity, we sample the LLM actions at a high temperature $T=0.8$ during simulation.
    \item \textbf{Backpropagation:} Evaluate the completed terminal report using the multi-dimensional reward function (described below) and propagate the reward upward to update the statistics of all ancestor nodes.
\end{enumerate}

\subsection{BERTScore-based Similarity Matrix for Keypoint Coverage}
\label{subsec:BERTScore}
We describe in detail how the BERTScore-based similarity matrix is constructed for the \textbf{Grounded Insight Verification --- Keypoint Coverage (GIV$_{\text{cov}}$)} metric.

\subsubsection{Background}
GIV$_{\text{cov}}$ measures the degree to which the sentences of a generated report cover the reference keypoints $K = \{k_1,\dots,k_M\}$. The metric is based on \textbf{normalized mutual information} (NMI) and thus requires a joint probability distribution, which we derive from a semantic similarity matrix between keypoint--sentence pairs.

\subsubsection{Constructing the similarity matrix $\mathbf{S}$}
Let $k_i$ be the $i$-th reference keypoint ($1 \leq i \leq M$) and $s_j$ be the $j$-th sentence (or sentence cluster) extracted from the generated report ($1 \leq j \leq N$).

For every pair $(k_i, s_j)$, we compute a similarity score $\mathbf{S}_{ij} \in [0,1]$ using \textbf{BERTScore}. The procedure consists of two steps:
\begin{enumerate}
    \item \textbf{Embedding.} A pre-trained BERT model (e.g., \texttt{bert-base-uncased} or \texttt{roberta-large}) is used to obtain token-level contextual embeddings for both $k_i$ and $s_j$.
    \item \textbf{Similarity computation.} BERTScore performs greedy cosine-similarity matching between the tokens of the source (keypoint) and the target (sentence). Let $\mathbf{x}_t$ be the embedding of the $t$-th token in the keypoint and $\mathbf{y}_u$ the embedding of the $u$-th token in the sentence. The token-level match score is
    \begin{equation}
        \mathrm{sim}_{\mathrm{token}}(t) = \max_u \cos(\mathbf{x}_t, \mathbf{y}_u).
    \end{equation}
    The sentence-level F1 score is then obtained by averaging the bidirectional matches (keypoint $\to$ sentence and sentence $\to$ keypoint) and applying a simple rescaling operation:
    \begin{equation}
        \mathbf{S}_{ij} = \mathrm{BERTScore}_{\mathrm{F1}}(k_i, s_j).
    \end{equation}
\end{enumerate}
Thus, $\mathbf{S}$ is an $M \times N$ matrix whose entries represent the semantic alignment between the corresponding keypoint--sentence pairs.

\subsubsection{Deriving probability distributions from the similarity matrix}
To transform the similarity matrix into quantities suitable for information-theoretic computation, we normalize it globally.

\begin{itemize}
    \item \textbf{Joint probability.} Normalizing each element by the total sum yields the joint probability distribution:
    \begin{equation}
        P(k_i, s_j) = \frac{\mathbf{S}_{ij}}{\sum_{i=1}^{M}\sum_{j=1}^{N} \mathbf{S}_{ij}}.
    \end{equation}
    This value represents the proportion of total semantic association that is captured by the pair $(k_i, s_j)$.

    \item \textbf{Marginal probabilities.} Summing over the appropriate dimensions gives the marginal probabilities for keypoints and sentences:
    \begin{equation}
        P(k_i) = \sum_{j=1}^{N} P(k_i, s_j), \qquad
        P(s_j) = \sum_{i=1}^{M} P(k_i, s_j).
    \end{equation}
    $P(k_i)$ reflects the relative semantic importance of keypoint $i$, while $P(s_j)$ captures the contribution of sentence $j$.
\end{itemize}

\subsubsection{Example}
Suppose there are $M=2$ keypoints and $N=3$ sentences, and the BERTScore matrix is
\[
\mathbf{S} =
\begin{bmatrix}
0.85 & 0.40 & 0.10 \\
0.20 & 0.90 & 0.30
\end{bmatrix}.
\]
The total sum is $0.85 + 0.40 + 0.10 + 0.20 + 0.90 + 0.30 = 2.75$. Hence,
\[
P(k_1,s_1) = \frac{0.85}{2.75} \approx 0.309,\quad
P(k_1) = 0.309+0.145+0.036 = 0.490,
\]
\[
P(s_1) = 0.309+0.073 = 0.382,\quad \text{etc.}
\]
These probabilities are then used directly in the NMI formula:
\begin{equation}
    \mathrm{GIV}_{\mathrm{cov}} = \frac{\sum_{i=1}^{M}\sum_{j=1}^{N} P(k_i,s_j) \log\frac{P(k_i,s_j)}{P(k_i)\,P(s_j)}}{-\sum_{i=1}^{M} P(k_i) \log P(k_i)}.
\end{equation}

\subsubsection{Design choices and remarks}
\begin{itemize}
    \item \textbf{Why BERTScore.} Unlike strict $n$-gram matching, BERTScore tolerates paraphrasing and only captures semantic equivalence --- a necessary property when evaluating keypoints that may be expressed in different wording.
    \item \textbf{Global normalization.} Transforming $\mathbf{S}$ into a probability distribution ensures that the NMI metric is well-behaved: it increases monotonically with coverage and approaches $1$ when all keypoints are fully covered and the report contains little irrelevant content.
    \item \textbf{Sentence clustering.} To improve stability, we first cluster sentences in the report and then compute similarity using the representative embedding of each cluster, mitigating noise from sentence-splitting differences.
\end{itemize}

This pipeline, which converts a BERTScore similarity matrix into a mutual-information-based coverage score, provides a \textbf{computable, reproducible, and paraphrase‑robust} metric for evaluating how faithfully a generated report captures the essential insights specified by the reference keypoints.

\subsection{Cross-Modal Consistency Evaluation}
\label{app:cross}

\paragraph{Definition.}
Cross-Modal Consistency (CMC) measures whether the textual and visual components of a generated report support each other. CMC is treated as the fourth dimension of T-Judge. It contains two complementary sub-metrics:
\begin{itemize}
    \item \textbf{Cross-Modal Reference Consistency (CMRC)}: whether sentences that refer to charts are factually and semantically consistent with the referenced charts and the underlying source tables.
    \item \textbf{Narrative Alignment (NRA)}: whether each generated chart is explained by the text, whether the explanation matches the chart's visual encoding, and whether the chart supports the analytical claim of its section.
\end{itemize}
Both CMRC and NRA are normalized to \([0,100]\). The final CMC score is
\begin{equation}
\mathrm{CMC}=0.5\cdot\mathrm{CMRC}+0.5\cdot\mathrm{NRA}.
\label{eq:cmc}
\end{equation}

\paragraph{CMRC: Cross-Modal Reference Consistency.}
Let \(R=\{r_1,\ldots,r_P\}\) be the set of sentences in the generated report that explicitly or implicitly refer to a chart. For each reference sentence \(r_p\), CMRC identifies the target chart \(c_p\), then verifies the following four aspects:
\begin{enumerate}
    \item \textbf{Existence}: the referenced chart \(c_p\) actually exists in the report.
    \item \textbf{Numerical consistency}: any numerical value, percentage, rank, or aggregate mentioned in \(r_p\) agrees with the chart data and the source table.
    \item \textbf{Trend consistency}: any directional or comparative claim (increase, decrease, peak, trough, faster, slower, etc.) agrees with the chart.
    \item \textbf{Encoding and label consistency}: the sentence correctly describes the chart type, axes, legend, units, categories, and visual encoding.
\end{enumerate}
Each reference sentence receives a score \(s(r_p)\in\{0,0.5,1\}\):
\begin{itemize}
    \item \(1\): all four aspects pass;
    \item \(0.5\): the reference and main trend are correct, but there is a minor numerical, label, or encoding error;
    \item \(0\): the chart does not exist, the core claim contradicts the chart, or the sentence misattributes a chart to the wrong evidence.
\end{itemize}
CMRC is computed as
\begin{equation}
\mathrm{CMRC}=100\cdot\frac{1}{|R|}\sum_{p=1}^{|R|}s(r_p).
\label{eq:cmrc}
\end{equation}
If the report contains no sentence that refers to any chart, CMRC is set to \(0\), because a multimodal report is expected to connect its visuals to its narrative.

\paragraph{NRA: Narrative Alignment.}
Let \(C=\{c_1,\ldots,c_Q\}\) be the set of generated charts. For each chart \(c_q\), NRA checks:
\begin{enumerate}
    \item whether at least one sentence or caption explains \(c_q\);
    \item whether the explanation matches the chart type, axes, legend, units, and visual pattern;
    \item whether the explanation supports the analytical claim of the section in which \(c_q\) appears.
\end{enumerate}
Each chart receives a score \(t(c_q)\in\{0,0.5,1\}\):
\begin{itemize}
    \item \(1\): the chart is clearly explained and the explanation is fully consistent with the visual encoding and the section claim;
    \item \(0.5\): the chart is mentioned, but the explanation is incomplete, vague, or partially inconsistent;
    \item \(0\): the chart is not explained at all, or the explanation contradicts the chart.
\end{itemize}
NRA is computed as
\begin{equation}
\mathrm{NRA}=100\cdot\frac{1}{|C|}\sum_{q=1}^{|C|}t(c_q).
\label{eq:nra}
\end{equation}
If the report contains no chart, NRA is set to \(0\) under the multimodal setting.

\paragraph{Prompt template.}
The CMC prompt is as follows.

\begin{verbatim}
System:
You are a strict cross-modal consistency evaluator for multimodal
business reports. You receive:
(1) the source tables,
(2) the generated textual report,
(3) the generated charts with captions,
(4) the section headings,
(5) the extracted chart data when available.
Your task is to compute CMRC and NRA.
Do not reward fluent writing. Penalize any chart-text mismatch,
unexplained chart, or text that refers to a non-existent chart.
Return only valid JSON.

User:
[SOURCE TABLES]
{table_text_or_markdown}

[GENERATED REPORT]
{report_text}

[CHARTS]
Chart ID: C1
Caption: {caption_1}
Section: {section_1}
Image: {image_1}
Extracted data: {chart_data_1}

Chart ID: C2
Caption: {caption_2}
Section: {section_2}
Image: {image_2}
Extracted data: {chart_data_2}

[TASK]
1. Identify every sentence that explicitly or implicitly refers to a chart.
2. For each such sentence, find the target chart ID and score it as 0, 0.5, or 1
   using the CMRC rules.
3. For each chart, determine whether the text explains it and score it as
   0, 0.5, or 1 using the NRA rules.
4. Return JSON with the following schema.

{
  "cmrc": [
    {
      "sentence_id": "S3",
      "sentence": "...",
      "chart_id": "C1",
      "score": 1,
      "reason": "..."
    }
  ],
  "nra": [
    {
      "chart_id": "C1",
      "score": 1,
      "reason": "..."
    }
  ],
  "cmrc_score": 100.0,
  "nra_score": 100.0,
  "cmc_score": 100.0
}
\end{verbatim}

\paragraph{Decision rules and tolerances.}
The following rules are used by the T-Judge's cross-modal submodule and, when needed, by human adjudicators.
\begin{itemize}
    \item \textbf{Numerical tolerance.} For absolute values, a claim is considered consistent if it is within \(1\%\) of the chart value. For proportions, percentages, or rates, the tolerance is \(2\) percentage points. Example: a chart value of \(5.0\%\) is consistent with a textual claim of \(5.0\%\pm 2.0\) percentage points.
    \item \textbf{Trend direction.} The direction of change must match exactly. ``Increase'' vs.\ ``decrease'' is a hard error. ``Sharp increase'' vs.\ ``slight increase'' is treated as a partial mismatch unless the chart clearly supports the magnitude.
    \item \textbf{Rank and comparison.} If the text says ``A is higher than B'', the chart must show the same ordering. Reversed ordering is a hard error.
    \item \textbf{Labels and units.} Category names, axis labels, legends, and units must match after normalization. Acceptable synonyms and unit conversions are allowed, e.g., ``USD million'' vs.\ ``million USD'', or ``Q1'' vs.\ ``first quarter''.
    \item \textbf{Chart type.} If the text says ``bar chart'' but the chart is a line plot, this is a minor error if the analytical claim is still correct, and a hard error if the chart type is central to the claim.
    \item \textbf{Missing chart reference.} If the text refers to ``Figure 3'' or ``the chart'' but no such chart exists, CMRC for that sentence is \(0\).
    \item \textbf{Unexplained chart.} If a chart appears in the report but no sentence or caption explains it, NRA for that chart is \(0\).
    \item \textbf{Extraction confidence.} If the chart-to-table extraction confidence is below \(0.8\), the ground-truth comparison uses the source table and the original chart specification instead of the extracted image data. If neither is available, the chart is marked as ``uncertain'' and excluded from the denominator, and this exclusion is reported in the appendix.
\end{itemize}

\paragraph{Examples.}
\textbf{Example 1 (CMRC = 1, NRA = 1).}
Text: ``As shown in Figure 1, employment declined sharply in 2009.''\\
Chart 1: a line chart showing a sharp drop in 2009.\\
Source table: the 2009 employment value is lower than 2008 by a large margin.\\
Decision: the chart exists, the trend matches, and the magnitude is supported. CMRC = 1. The chart is explained by the text and supports the section claim. NRA = 1.

\textbf{Example 2 (CMRC = 0, NRA = 0.5).}
Text: ``Figure 2 shows a \(20\%\) growth in 2010.''\\
Chart 2: a bar chart showing \(5\%\) growth in 2010.\\
Decision: the chart exists and is mentioned, but the numerical claim is wrong. CMRC = 0. NRA = 0.5 because the chart is referenced but the explanation contradicts the visual evidence.

\paragraph{Reproducibility.}
The T-Judge's cross-modal submodule uses the same model snapshot as T-Judge and V-Judge: \texttt{gpt-4o}, temperature \(0.2\), top-\(p=1.0\), and a fixed random seed. Each report is evaluated three times, and the mean CMRC, NRA, and CMC are reported. The parsing failure rate is below \(1\%\); failed parses are retried up to three times and then manually adjudicated.

\subsection{Dual-Judge Evaluation: Implementation Details}
\label{app:judge}

The following provides an implementation-level completion of the \textbf{DUAL-JUDGE EVALUATION} section. The scoring protocol strictly follows the original text:
\begin{itemize}
    \item \textbf{T-Judge}: \(T=\mathrm{mean}(SC, GIV, NA, CMC)\), where \(GIV=\mathrm{mean}(GIV_{\mathrm{num}}, GIV_{\mathrm{cov}})\) and \(CMC=0.5\cdot CMRC+0.5\cdot NRA\).
    \item \textbf{V-Judge}: \(V=\mathrm{mean}(DF, EX, AE, TC)\).
    \item \textbf{Overall}: \(0.6T+0.4V\). All dimensions and the overall score are normalized to \([0,100]\).
    \item Unified judge settings: GPT-4o / GPT-4o-vision, temperature \(0.2\), top-\(p=1.0\), fixed random seed. Each report is evaluated three times, and the mean is reported. Parse failures are retried up to three times and then manually adjudicated.
\end{itemize}

The original paper only provides the full CMC prompt; the remaining prompts below are completed under the same JSON-output protocol and can be used directly as appendix implementations.

\subsubsection{T-Judge Dimensions}

\paragraph{Structural Completeness (SC)}

\textbf{Implementation.}
\begin{itemize}
    \item Heading matching: compute semantic embedding matching between generated headings and reference headings with threshold \(0.85\), and compute F1 as \(SC_{\mathrm{head}}\).
    \item Content coverage: T-Judge determines whether matched sections cover the same analytical sub-dimensions, yielding \(SC_{\mathrm{content}}\in[0,100]\).
    \item Aggregation:
    \[
    SC=0.3\cdot SC_{\mathrm{head}}+0.7\cdot SC_{\mathrm{content}}.
    \]
\end{itemize}

\textbf{SC prompt snippet.}
\begin{verbatim}
SC task:
Given reference chapter structure and generated report, compute:
- SC_head: heading F1 with semantic equivalence threshold 0.85.
- SC_content: 0-100, whether matched sections cover the same analytical sub-dimensions.
Return SC_head, SC_content, SC = 0.3*SC_head + 0.7*SC_content.
\end{verbatim}

\paragraph{Grounded Insight Verification (GIV)}

\paragraph{(a) Numerical Accuracy, \(GIV_{\mathrm{num}}\).}
\textbf{Implementation.}
\begin{enumerate}
    \item Extract numerical sentence clusters.
    \item Generate verification questions for each numerical claim, with the claimed value as the target answer.
    \item Have three specialized code-generation LLMs independently generate and execute SQL queries on the source tables.
    \item A claim is verified if at least \(2/3\) models agree within a \(1\%\) tolerance margin.
    \item \(GIV_{\mathrm{num}}\) is the proportion of verified claims multiplied by \(100\).
\end{enumerate}

\textbf{Numerical verification question generation prompt.}
\begin{verbatim*}
System:
You are a numerical verification question generator for table-grounded reports.
Return valid JSON only.

User:
[SOURCE TABLES]
{table_text_or_markdown}

[GENERATED REPORT]
{report_text}

[TASK]
1. Extract numerical sentence clusters.
2. For each claimed value, generate a verification question whose target answer is the claimed value.
3. Generate an executable SQL query over the source tables.
4. Return JSON:
{
  "claims": [
    {
      "claim_id": "N1",
      "sentence": "...",
      "claimed_value": ...,
      "verification_question": "...",
      "sql": "..."
    }
  ]
}
\end{verbatim*}

\paragraph{(b) Keypoint Coverage, \(GIV_{\mathrm{cov}}\).}
\textbf{Implementation.}
\begin{itemize}
    \item Use BERTScore to compute the similarity matrix \(S_{ij}\) between reference keypoints \(k_i\) and generated sentences \(s_j\).
    \item Normalize globally to obtain the joint probability \(P(k_i,s_j)\), and then obtain the marginals \(P(k_i)\) and \(P(s_j)\).
    \item Compute coverage using normalized mutual information:
    \[
    GIV_{\mathrm{cov}}=
    \frac{\sum_{i=1}^{M}\sum_{j=1}^{N}P(k_i,s_j)\log\frac{P(k_i,s_j)}{P(k_i)P(s_j)}}{-\sum_{i=1}^{M}P(k_i)\log P(k_i)}.
    \]
    \item This dimension requires no additional LLM prompt and is a reproducible semantic matching metric.
\end{itemize}

\textbf{GIV aggregation.}
\[
GIV=\frac{GIV_{\mathrm{num}}+GIV_{\mathrm{cov}}}{2}.
\]

\paragraph{ Novelty Analysis (NA)}

\textbf{Implementation.}
\begin{itemize}
    \item First identify novel claims: sentence clusters whose maximum cosine similarity to any reference keypoint is below \(\tau_{\mathrm{novel}}=0.6\).
    \item T-Judge scores each novel claim on three sub-dimensions: non-triviality, actionability, and correctness.
    \item Let each claim receive a normalized score \(c_n\in[0,1]\), and let \(|N|\) be the number of novel claims.
    \item Penalty factor:
    \[
    \min\left(1,\frac{5}{|N|}\right).
    \]
    \item If \(|N|=0\), then \(NA=0\); otherwise:
    \[
    NA=100\cdot \frac{1}{|N|}\sum_{n=1}^{|N|}c_n\cdot \min\left(1,\frac{5}{|N|}\right).
    \]
\end{itemize}

\textbf{NA prompt snippet.}
\begin{verbatim}
NA task:
A novel claim is a sentence cluster whose maximum cosine similarity to any reference keypoint is below 0.6.
For each novel claim, score:
- non-triviality: 0-5
- actionability: 0-5
- correctness: factual consistency with source tables, 0-5
Normalize each claim to [0,1]. Apply penalty min(1, 5/|N|).
If |N|=0, NA=0.
Return novel_claims, per-claim score, NA.
\end{verbatim}

\paragraph{Cross-Modal Consistency (CMC)}

\textbf{Implementation.}
\[
CMC=0.5\cdot CMRC+0.5\cdot NRA.
\]
\begin{itemize}
    \item \textbf{CMRC}: For each sentence that explicitly or implicitly refers to a chart, check existence, numerical consistency, trend consistency, and encoding/label consistency. The score is \(s(r_p)\in\{0,0.5,1\}\).
    \[
    CMRC=100\cdot\frac{1}{|R|}\sum_{p=1}^{|R|}s(r_p).
    \]
    If there is no chart-referencing sentence, \(CMRC=0\).
    \item \textbf{NRA}: For each chart, check whether it is explained by the text, whether the explanation matches the chart encoding, and whether it supports the analytical claim of the section. The score is \(t(c_q)\in\{0,0.5,1\}\).
    \[
    NRA=100\cdot\frac{1}{|C|}\sum_{q=1}^{|C|}t(c_q).
    \]
    If there is no chart, \(NRA=0\).
\end{itemize}

\textbf{CMC full prompt.}
\begin{verbatim*}
System:
You are a strict cross-modal consistency evaluator for multimodal business reports.
You receive:
(1) source tables,
(2) generated textual report,
(3) generated charts with captions,
(4) section headings,
(5) extracted chart data when available.
Your task is to compute CMRC and NRA.
Do not reward fluent writing.
Penalize any chart-text mismatch, unexplained chart, or text that refers to a non-existent chart.
Return only valid JSON.

User:
[SOURCE TABLES]
{table_text_or_markdown}

[GENERATED REPORT]
{report_text}

[CHARTS]
Chart ID: C1
Caption: {caption_1}
Section: {section_1}
Image: {image_1}
Extracted data: {chart_data_1}

Chart ID: C2
Caption: {caption_2}
Section: {section_2}
Image: {image_2}
Extracted data: {chart_data_2}

[TASK]
1. Identify every sentence that explicitly or implicitly refers to a chart.
2. For each such sentence, find the target chart ID and score it as 0, 0.5, or 1 using CMRC rules.
3. For each chart, determine whether the text explains it and score it as 0, 0.5, or 1 using NRA rules.
4. Return JSON:
{
  "cmrc": [
    {
      "sentence_id": "S3",
      "sentence": "...",
      "chart_id": "C1",
      "score": 1,
      "reason": "..."
    }
  ],
  "nra": [
    {
      "chart_id": "C1",
      "score": 1,
      "reason": "..."
    }
  ],
  "cmrc_score": 100.0,
  "nra_score": 100.0,
  "cmc_score": 100.0
}
\end{verbatim*}

\textbf{Decision rules and tolerances.}
\begin{itemize}
    \item Absolute values: \(1\%\) tolerance; proportions, percentages, or rates: \(2\) percentage points.
    \item Trend direction must match exactly; reversing order is a hard error.
    \item Labels, units, and chart type follow the original rules.
    \item If chart-to-table extraction confidence is below \(0.8\), use the source table and the original chart specification. If neither is available, mark the chart as uncertain and exclude it from the denominator.
\end{itemize}

\subsubsection{V-Judge Dimensions}

\paragraph{Data Fidelity (DF)}

\textbf{Implementation.}
\begin{enumerate}
    \item A chart-to-table model extracts data from the chart.
    \item SQL generates expected data from the source tables.
    \item Comparison: if more than \(5\%\) of data points deviate beyond a \(1\%\) absolute tolerance or a \(2\) percentage-point tolerance for proportions, the chart is flagged as a fidelity failure.
    \item DF is aggregated as a chart-level data-fidelity score, normalized to \([0,100]\).
\end{enumerate}

\paragraph{Expressiveness (EX), Aesthetics (AE), and Technical Correctness (TC)}

\textbf{Implementation.}
\begin{itemize}
    \item Each chart is scored on a \(0\)--\(5\) scale, then multiplied by \(20\) to convert to \([0,100]\).
    \item EX: chart type appropriateness, encoding clarity, and insight salience.
    \item AE: color harmony, layout quality, and absence of visual noise, with calibration examples.
    \item TC: rule-based checks for missing elements, scale errors, zero-baseline checks, label readability, and color accessibility.
\end{itemize}

\textbf{V-Judge full prompt.}
\begin{verbatim*}
System:
You are a strict V-Judge for generated visualizations in multimodal business reports.
Evaluate only against source tables, chart specifications, and section claims.
Return valid JSON only.

User:
[SOURCE TABLES]
{table_text_or_markdown}

[CHARTS]
Chart ID: C1
Caption: {caption_1}
Section: {section_1}
Image: {image_1}
Extracted data: {chart_data_1}
Extraction confidence: {confidence_1}
Expected data from SQL: {expected_data_1}

[TASK]
For each chart:
- DF: compare extracted data with expected data. Mark fidelity failure if >5% data points deviate beyond 1% absolute tolerance or 2 percentage points for proportions. Return DF 0-100.
- EX: 0-5, chart type appropriateness, encoding clarity, insight salience.
- AE: 0-5, color harmony, layout quality, absence of visual noise.
- TC: 0-5, missing elements, scale errors including zero-baseline, label readability, color accessibility.
Return JSON:
{
  "charts": [
    {
      "chart_id": "C1",
      "DF": 100.0,
      "EX": 4,
      "AE": 4,
      "TC": 5,
      "fidelity_failure": false,
      "reason": "..."
    }
  ],
  "DF": 100.0,
  "EX": 80.0,
  "AE": 80.0,
  "TC": 100.0,
  "V_score": 90.0
}
\end{verbatim*}

\textbf{V-Judge aggregation.}
\[
V=\frac{DF+EX+AE+TC}{4},
\]
where EX, AE, and TC have already been mapped from \(0\)--\(5\) to \(0\)--\(100\).

\subsubsection{Final Scoring}

\[
T=\frac{SC+GIV+NA+CMC}{4},
\]
\[
V=\frac{DF+EX+AE+TC}{4},
\]
\[
Overall=0.6\cdot T+0.4\cdot V.
\]

All dimensions and the overall score are normalized to \([0,100]\). This implementation is consistent with the original text: T-Judge handles text, factuality, novelty, and cross-modal consistency; V-Judge handles chart data fidelity, expressiveness, aesthetics, and technical correctness. The dual-judge separation avoids the \(8\)--\(12\) point inflation of a single judge and prevents cross-modal compensation.

\subsection{Ablation: Sensitivity to Judge Weights.}
We vary the \TJudge–\VJudge weight ratio from (0.5,0.5) to (0.8,0.2). The relative ordering of top models remains stable, with DeepSeek-R1 leading under all settings. The largest ranking shift occurs in the mid-tier: code-augmented models gain under higher \VJudge weight, while deep research agents, which often produce richer text but less polished charts, benefit from higher \TJudge weight. The default (0.6,0.4) balances these tendencies.

\subsection{Human Evaluation and Consistency}

\begin{table}[t]
\centering
\caption{Pearson correlation between automated \bench{} overall scores and human expert judgments, broken down by dimension.}
\label{tab:human_corr}
\begin{tabular}{p{6cm} c}
\toprule
\textbf{Dimension} & \textbf{Pearson} $r$ \\
\midrule
Structural Completeness (SC) & 0.88 \\
Grounded Insight Verification (GIV) & 0.94 \\
Novelty Analysis (NA) & 0.82 \\
Data Fidelity (DF) & 0.91 \\
Expressiveness (EX) & 0.85 \\
Aesthetics (AE) & 0.79 \\
Technical Correctness (TC) & 0.90 \\
Overall Score & 0.91 \\
\bottomrule
\end{tabular}
\end{table}

\subsection{Error Analysis}
\label{subsec:error}

\begin{figure*}[t]
\centering
\includegraphics[width=1\linewidth]{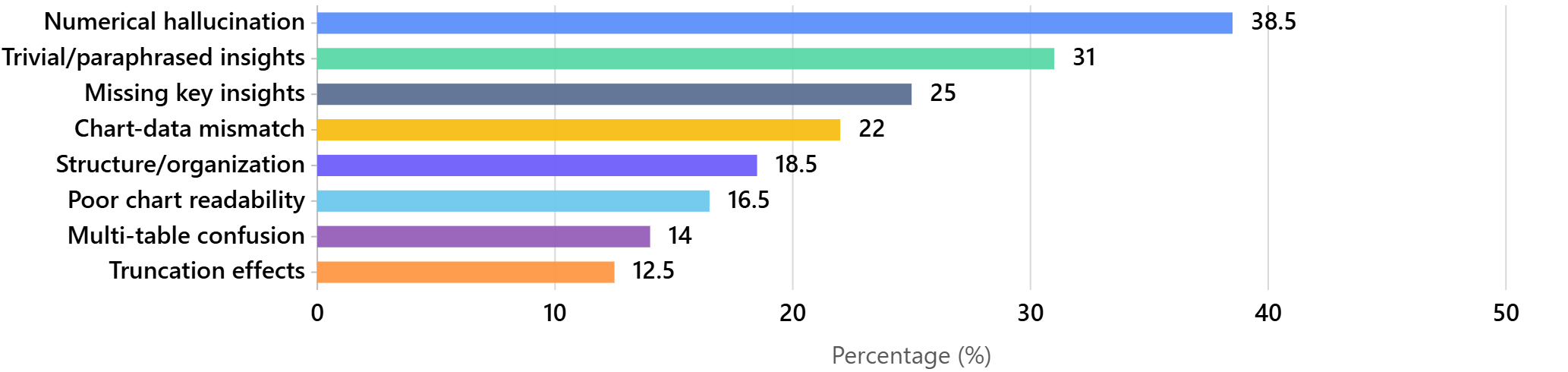}
\caption{The frequency of common error categories across 200 sampled generated reports. Numerical hallucination is the most pervasive issue (38.5\%), followed by trivial/paraphrased insights (31.0\%) and missing key insights (25.0\%). The distribution pinpoints factual accuracy, analytical depth, and multi‑table reasoning as the primary weaknesses of current models.}
\label{fig:error}
\end{figure*}

We manually analyze 200 randomly sampled reports to categorize failure modes. The error taxonomy is extended for multimodal settings. Figure~\ref{fig:error} shows the distribution of primary error types in generated report.

\begin{table}[t]
\centering
\caption{Distribution of primary error types in generated reports. A single report may contain multiple errors.}
\label{tab:error_types}
\begin{tabular}{lcc}
\toprule
\textbf{Error Type} & \textbf{Frequency (\%)} & \textbf{Metric(s)} \\
\midrule
Numerical hallucination & 38.5\% & GIV$_{\text{num}}$, DF \\
Missing key insights & 25.0\% & GIV$_{\text{cov}}$ \\
Structure/organization issues & 18.5\% & SC \\
Chart-data mismatch & 22.0\% & DF \\
Poor chart readability & 16.5\% & EX, AE, TC \\
Trivial/paraphrased insights & 31.0\% & NA \\
Multi-table confusion & 14.0\% & GIV, DF \\
Truncation effects (large tables) & 12.5\% & GIV$_{\text{num}}$ \\
\bottomrule
\end{tabular}
\end{table}

\subsection{Case Study}
\label{sec:case_study}

\begin{figure*}[t]
\centering
\includegraphics[width=1\linewidth]{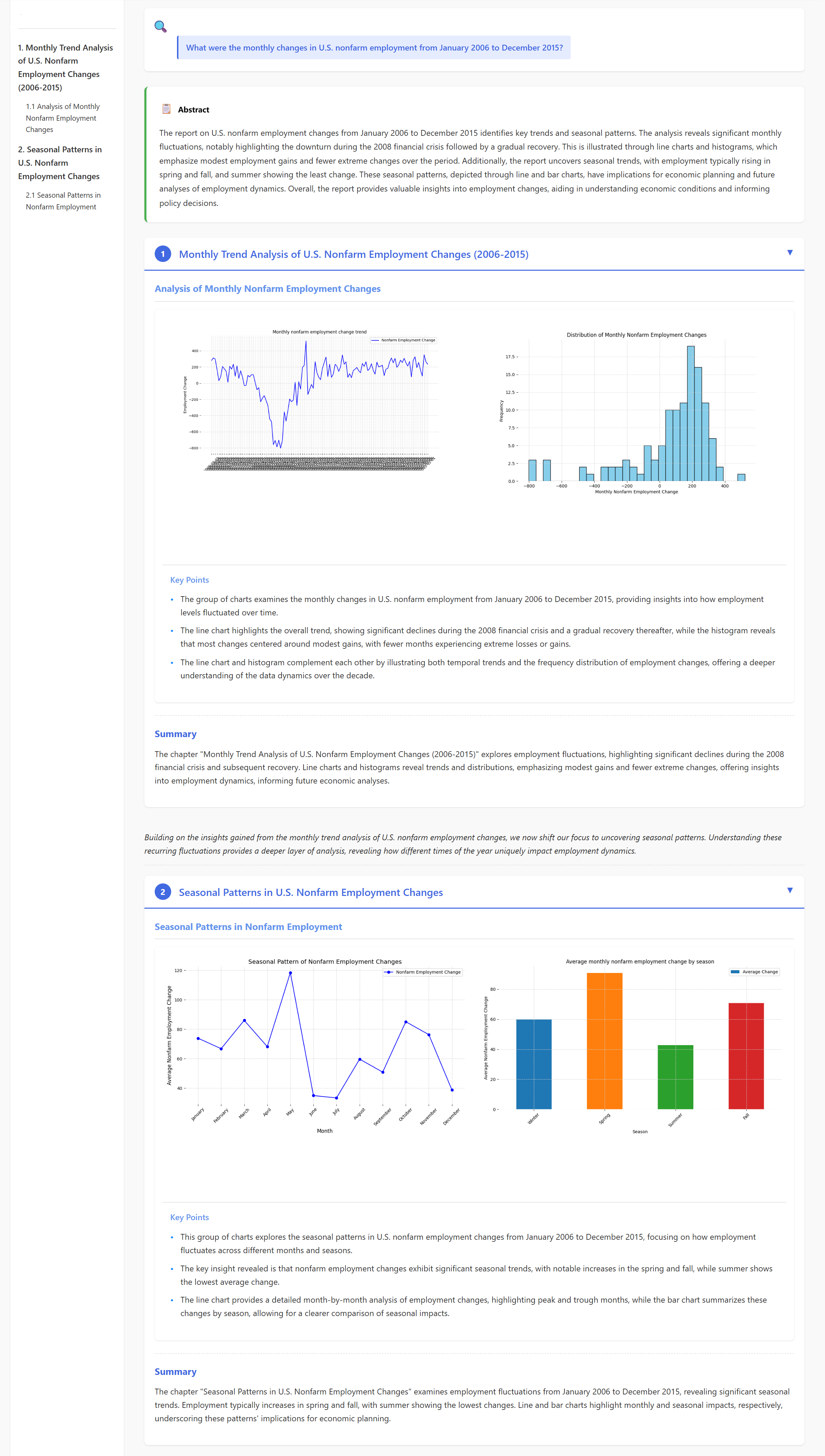}
\caption{An example report generated by GPT-4o.}
\label{fig:example}
\end{figure*}

Using the U.S. employment trends report as an illustrative case (Figure~\ref{fig:example}), our dual-judge evaluation reveals a mixed performance profile: the report achieves strong structural completeness (SC 92) through its well‑organized chapter hierarchy, but its grounded insight verification is only moderate (GIV 65) because many statements remain qualitative without explicit numerical anchoring, and novelty is low (NA 30) as the insights largely paraphrase visible trends rather than offering non‑trivial causal analysis or actionable recommendations. The accompanying charts are functionally correct but suffer from missing annotations and cluttered legends, yielding a VisScore of 79. The overall \bench score of 69.3, while higher than the best model‑generated reports (which typically score in the low 60s), still falls well short of the human baseline, demonstrating that even a well‑structured and factually faithful report can be penalized for insufficient quantitative grounding and a lack of deep analytical insight—precisely the weaknesses that our benchmark is designed to expose.

\newpage
\end{document}